\documentclass{article}

\PassOptionsToPackage{numbers}{natbib}

\usepackage[final]{neurips_2022}

\usepackage[english]{babel}
\usepackage{amsthm}

\usepackage{CJK}

\theoremstyle{definition}
\newtheorem{definition}{Definition}

\newtheorem{lemma}{Lemma}
\newtheorem{theorem}{Theorem}

\newtheorem{assumption}{Assumption}

\theoremstyle{remark}

\usepackage[utf8]{inputenc} 
\usepackage[T1]{fontenc}    
\usepackage[breaklinks=true,colorlinks,bookmarks=false]{hyperref}
\usepackage{url}            
\usepackage{booktabs}       
\usepackage{amsfonts}       
\usepackage{nicefrac}       
\usepackage{microtype}      
\usepackage{xcolor}         
\usepackage{mathtools}
\usepackage{mathrsfs}
\usepackage{tabularx}
\usepackage{verbatim}
\usepackage{graphicx} 
\usepackage{subfigure}
\usepackage{wrapfig}
\usepackage{algorithm, algpseudocode}
\newcommand{\yinpeng}[1]{\textcolor{red}{[yp:#1]}}

\usepackage{graphicx}
\usepackage{float} 
\usepackage{subfigure}
\usepackage[]{caption2} 
\renewcommand{\thesubfigure}{(\alph{subfigure})}
\makeatletter \renewcommand{\@thesubfigure}{\thesubfigure \space}
\renewcommand{\p@subfigure}{} \makeatother
\newcommand{\vect}[1]{\boldsymbol{\mathbf{#1}}}

\DeclareMathOperator*{\argmax}{arg\,max} 

\title{Pre-trained Adversarial Perturbations}

\renewcommand\footnotemark{}
\author{%
     Yuanhao Ban$^{1,2*}$, Yinpeng Dong$^{1,3\dagger}$\thanks{$^*$This work was done when Yuanhao Ban was intern at RealAI, Inc; $^{\dagger}$Corresponding author.}\\
     $^{1}$ Department of Computer Science \& Technology, Institute for AI, BNRist Center,\\
     Tsinghua-Bosch Joint ML Center, THBI Lab, Tsinghua University \\
     $^{2}$ Department of Electronic Engineering, Tsinghua University  \hspace{2ex}  $^{3}$ RealAI\\
     \footnotesize{\texttt{banyh19@mails.tsinghua.edu.cn, dongyinpeng@mail.tsinghua.edu.cn}}
}

\begin{document}

\maketitle

\begin{abstract}
Self-supervised pre-training has drawn increasing attention in recent years due to its superior performance on numerous downstream tasks after fine-tuning. However, it is well-known that deep learning models lack the robustness to adversarial examples, which can also invoke security issues to pre-trained models, despite being less explored. In this paper, we delve into the robustness of pre-trained models by introducing Pre-trained Adversarial Perturbations (PAPs), which are universal perturbations crafted for the pre-trained models to maintain the effectiveness when attacking fine-tuned ones without any knowledge of the downstream tasks. To this end, we propose a Low-Level Layer Lifting Attack (L4A) method to generate effective PAPs by lifting the neuron activations of low-level layers of the pre-trained models. Equipped with an enhanced noise augmentation strategy, L4A is effective at generating more transferable PAPs against fine-tuned models. Extensive experiments on typical pre-trained vision models and ten downstream tasks demonstrate that our method improves the attack success rate by a large margin compared with state-of-the-art methods.

\end{abstract}

\section{Introduction}\label{sec:1}

Large-scale pre-trained models~\cite{qiu2020pre,han2021pre} have recently achieved unprecedented success in a variety of fields, e.g., natural language processing~\cite{devlin2018bert,liu2019roberta,brown2020language}, computer vision~\cite{chen2020simple,he2020momentum,he2021masked}. A large amount of work proposes sophisticated self-supervised learning algorithms, enabling the pre-trained models to extract useful knowledge from large-scale unlabeled datasets. The pre-trained models consequently facilitate downstream tasks through transfer learning or fine-tuning~\cite{oquab2014transfer,yosinski2014transferable,guo2019spottune}. Nowadays, more practitioners without sufficient computational resources or training data tend to fine-tune the publicly available pre-trained models on their own datasets. Therefore, it has become an emerging trend to adopt the paradigm of pre-training to fine-tuning rather than training from scratch~\cite{han2021pre}.


Despite the excellent performance of deep learning models, they are incredibly vulnerable to adversarial examples~\cite{szegedy2013FGSM,goodfellow2014adver}, which are generated by adding small, human-imperceptible perturbations to natural examples, but can make the target model output erroneous predictions. Adversarial examples also exhibit an intriguing property called \emph{transferability}~\cite{szegedy2013FGSM,liu2016blackbox,moosavi2017universal}, which means that the adversarial perturbations generated for one model or a set of images can remain adversarial for others. For example, a universal adversarial perturbation (UAP)~\cite{moosavi2017universal} can be generated for the entire distribution of data samples, demonstrating excellent cross-data transferability. Other work~\cite{liu2016blackbox,dong2018boosting,xie2019improving,dong2019evading,naseer2019cross} has revealed that adversarial examples have high cross-model and cross-domain transferability, making \emph{black-box attacks} practical without any knowledge of the target model or even the training data. However, much less effort has been devoted to exploring the adversarial robustness of pre-trained models. As these models have been broadly studied and deployed in various real-world applications, it is of significant importance to identify their weaknesses and evaluate their robustness, especially concerning the pre-training to the fine-tuning procedure.

\begin{figure}
\centering
\includegraphics[width=0.95\linewidth]{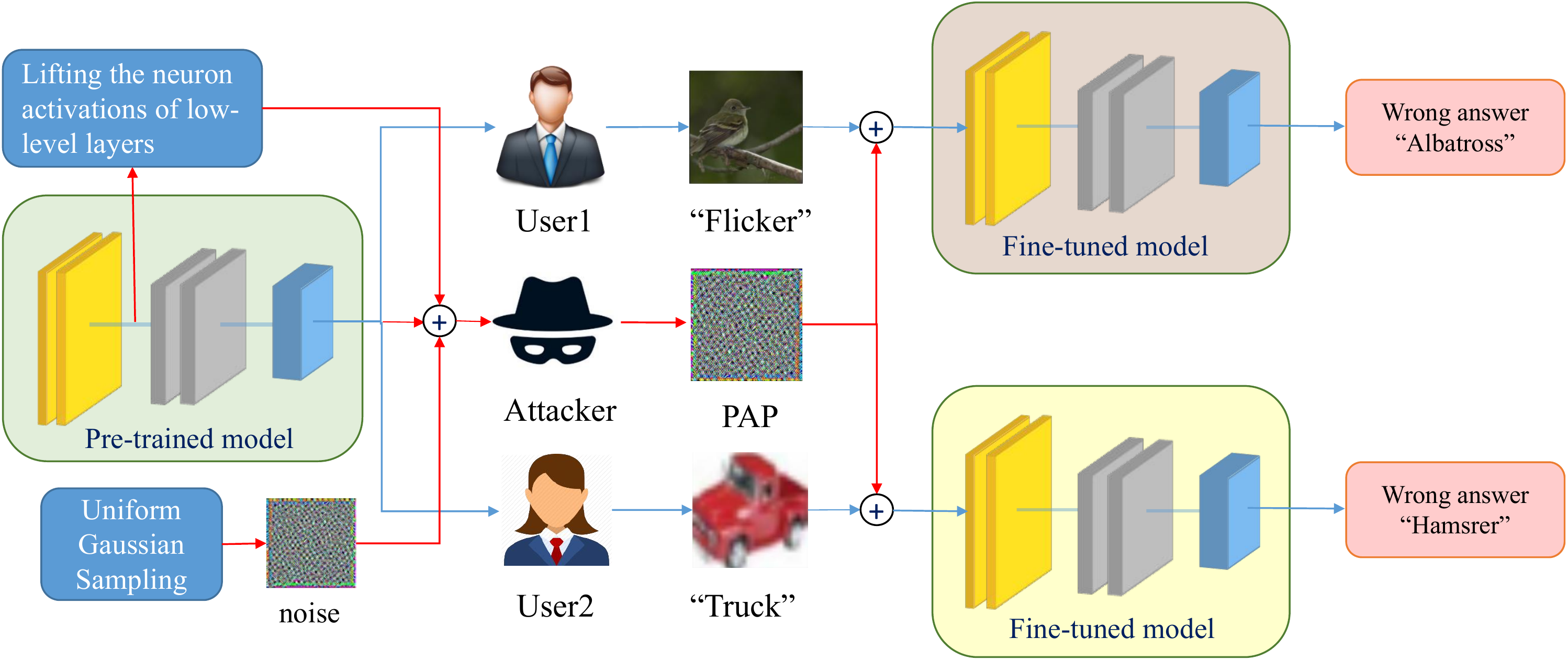}
\caption{A demonstration of pre-trained adversarial perturbations (PAPs): An attacker first downloads pre-trained weights on the Internet and generates a PAP by lifting the neuron activations of low-level layers of the pre-trained models. We adopt a data augmentation technique called uniform Gaussian sampling to improve the transferability of PAP. When users fine-tune the pre-trained models to complete downstream tasks, the attacker can add the PAP to the input of the fine-tuned models to cheat them without knowing the specific downstream tasks.}
\label{fig9}
\end{figure}

In this paper, we introduce \textbf{Pre-trained Adversarial Perturbations (PAPs)}, a new kind of universal adversarial perturbations designed for pre-trained models. Specifically, a PAP is generated for a pre-trained model to effectively fool any downstream model obtained by fine-tuning the pre-trained one, as illustrated in Fig.~\ref{fig9}. It works under a quasi-black-box setting where the downstream task, dataset, and fine-tuned model parameters are all unavailable. This attack setting is more suitable for the pre-training to the fine-tuning procedure since many pre-trained models are publicly available, and the adversary may generate PAPs before the pre-trained model has been fine-tuned. Although there are many methods~\cite{dong2018boosting,xie2019improving} proposed for improving the transferability, they do not consider the specific characteristics of the pre-training to the fine-tuning procedure, limiting their \emph{cross-finetuning transferability} in our setting.


To generate more effective PAPs, we propose a \textbf{Low-Level Layer Lifting Attack (L4A)} method, which aims to lift the feature activations of low-level layers. Motivated by the finding that the lower the level of the model's layer is, the less its parameters change during fine-tuning, we generate PAPs to destroy the low-level
feature representations of pre-trained models, making the attacking effects better reserved after fine-tuning. To further alleviate the overfitting of PAPs to the source domain, we improve L4A with a noise augmentation technique. We conduct extensive experiments on typical pre-trained vision models~\cite{chen2020simple,he2021masked} and ten downstream tasks. The evaluation results demonstrate that our method achieves a higher attack success rate on average compared with the alternative baselines. 



\section{Related work}\label{sec:related work}

\textbf{Self-supervised learning.}
Self-supervised learning (SSL) enables learning from unlabeled data. To achieve this, early approaches utilize hand-crafted pretext tasks, including colorization~\cite{2016Colorful}, rotation prediction~\cite{gidaris2018unsupervisedrotation}, position prediction~\cite{noroozi2016jig}, and Selfie~\cite{trinh2019selfie}. Another approach for SSL is contrastive learning~\cite{le2020contrastive,park2020contrastive,chen2020simple,khosla2020supervised}, which aims to map the input image to the feature space and minimize the distance between similar ones while keeping dissimilar ones far away from each other. In particular, a similar sample is retrieved by applying appropriate data augmentation techniques to the original one, and the versions of different samples are viewed as dissimilar pairs.

\textbf{Adversarial examples.}
With the knowledge of the structure and parameters of a model, many algorithms~\cite{kurakin2018physical,moosavi2016deepfool,madry2017PGD,papernot2016lJMSA} successfully fool the target model in a white-box manner. 
An intriguing property of adversarial examples is their good transferability~\cite{liu2016blackbox,moosavi2017universal}. The universal adversarial perturbations~\cite{moosavi2017universal} demonstrate good cross-data transferability by optimizing under a distribution of data samples. The cross-model transferability has also been extensively studied~\cite{dong2018boosting,xie2019improving,dong2019evading}, enabling the attack on black-box models without any knowledge of their internal working mechanisms. 

\textbf{Robustness of the pre-training to fine-tuned procedure.}
Due to the popularity of pre-trained models, a lot of works~\cite{sai2020improving,yang2022robust,chen2021towards} study the robustness of this setting. Among them, Dong et al.~\cite{dong2021should} propose a novel adversarial fine-tuning method in an information-theoretical way to retain robust features learned from the pre-trained model. Jiang et al.~\cite{jiang2020robust} integrate adversarial samples into the pre-training procedure to defend against attacks. Fan et at.~\cite{fan2021does} adopt Clusterfit~\cite{yan2020clusterfit} to generate pseudo-label data and later use them for training the model in a supervised way, which improves the robustness of the pre-trained model. The main difference between our work and theirs is that we consider the problem from an attacker's perspective.

\section{Methodology}\label{sec:method}

In this section, we first introduce the notations and the problem formulation of the Pre-trained Adversarial Perturbations (PAPs). Then, we detail the Low-Level Layer Lifting Attack (L4A) method. 

\subsection{Notations and problem formulation}
Let $f_{\vect{\theta}}$ denote a pre-trained model for feature extraction with parameters $\vect{\theta}$. It takes an image ${\vect{x}}$ $\in$ $\mathcal{D}_p$ as input and outputs a feature vector ${\vect{v}}$ $\in$ $\mathcal{X}$, where $\mathcal{D}_p$ and $\mathcal{X}$ refer to the pre-training dataset and feature space, respectively. 
We denote $f_{\vect{\theta}}^{k}({\vect{x}})$ as the $k$-th layer's feature map of $f_{\vect{\theta}}$ for an input image ${\vect{x}}$. In the pre-training to fine-tuning paradigm, a user fine-tunes the pre-trained model $f_{\vect{\theta}}$ using a new dataset $\mathcal{D}_t$ of the downstream task and finally gets a fine-tuned model $f_{\vect{\theta'}}$ with updated parameters ${\vect{\theta'}}$. Then, let $f_{\vect{\theta'}}({\vect{x}})$ be the predicted probability distribution of an image $\vect{x}$ over the classes of $\mathcal{D}_t$, and $F_{\vect{\theta'}}({\vect{x}}) = \argmax f_{\vect{\theta'}}({\vect{x}})$ be the final classification result. 

In this paper, we introduce  \textbf{Pre-trained Adversarial Perturbations (PAPs)}, which are generated for the pre-trained model $f_{\vect{\theta}}$, but can effectively fool fine-tuned models $f_{\vect{\theta'}}$ on downstream tasks. Formally, a PAP is a universal perturbation $\vect{\delta}$ within a small budget $\epsilon$ crafted by $f_{\vect{\theta}}$ and $\mathcal{D}_p$, such that $F_{\vect{\theta'}}({\vect{x}}+\vect{\delta}) \neq F_{\vect{\theta'}}(\vect{x})$ for most of the instances belonging to the fine-tuning dataset $\mathcal{D}_t$. This can be formulated as the following optimization problem:
\begin{equation}
\label{optimization}
 \max_{\vect{\delta}}\mathbb{E}_{{\vect{x}}\sim \mathcal{D}_t}[F_{\vect{\theta'}}({\vect{x}})\neq F_{\vect{\theta'}}({\vect{x}}+\vect{\delta})], \; \text{s.t.}\;\|\vect{\delta}\|_{p}\leq \epsilon\;  \text{and }{\vect{x}}+\vect{\delta} \in \left [0,1\right],
\end{equation}
where $\|\cdot\|_p$ denotes the $\ell_p$ norm and we take the $\ell_{\infty}$ norm in this work.
There exist some works related to the universal perturbations, such as the universal adversarial perturbation (UAP)~\cite{moosavi2017universal} and the fast feature fool (FFF)~\cite{mopuri2017fff}, as detailed below.

\textbf{UAP}: Given a classifier $f$ and its dataset $\mathcal{D}$, the UAP tries to generate a perturbation $\vect{\delta}$ that can fool the model on most of the instances from $\mathcal{D}$, which is usually solved by an iterative method. Every time sampling an image ${\vect{x}}$ from the dataset $\mathcal{D}$, the attacker computes the minimal perturbation $\vect{\zeta}$ that sends ${\vect{x}}+\vect{\delta}$ to the decision boundary by Eq.~\eqref{eq:UAP} and then adds it into $\vect{\delta}$.
\begin{equation}
\label{eq:UAP}
\vect{\zeta} \leftarrow \arg\min_{\vect{r}}\|\vect{r}\|_2,  \;\text{s.t.}\;F(\vect{x}+\vect{\delta}+\vect{r}) \neq F(\vect{x}).
\end{equation}

\textbf{FFF}: It aims to produce maximal spurious activations at each layer. To achieve this, FFF starts with a random $\vect{\delta}$ and solves the following problem:
\begin{equation}
\label{eq:FFF}
\min_{\vect{\delta}} -\log\left(\prod \limits_{i=0}^K\overline{l}_i(\vect{\delta})\right),\;\text{s.t.}\;\|\vect{\delta}\|_{p}\leq \epsilon.
\end{equation}
where $\overline{l}_i(\vect{\delta})$ is the mean of the output tensor at layer $i$.
\subsection{Our design}
However, these attacks show limited cross-finetuning transferability in our problem setting due to ignorance of the fine-tuning procedure. Two challenges are degenerating the performance.  

\begin{itemize}
\item[$\bullet$] \textbf{Fine-tuning Deviation.} The parameters of the model could change a lot during fine-tuning. As a result, the generated adversarial samples may perform well in the feature space of the pre-trained model but fail in the fine-tuned ones.  
\item[$\bullet$] \textbf{Datasets Deviation.} The statistics (i.e., mean and standard deviation) of different datasets can vary a lot. Only using the pre-training dataset with the fixed statistics to generate adversarial samples may suffer a performance drop. 
\end{itemize}

To alleviate the negative effect of the above issues, we propose a \textbf{Low-Level Layer Lifting Attack (L4A)} method equipped with a \textbf{uniform Gaussian sampling} strategy.

\begin{figure}[t]
\centering  
\subfigure[Resnet50]{
\includegraphics[width=0.32\textwidth]{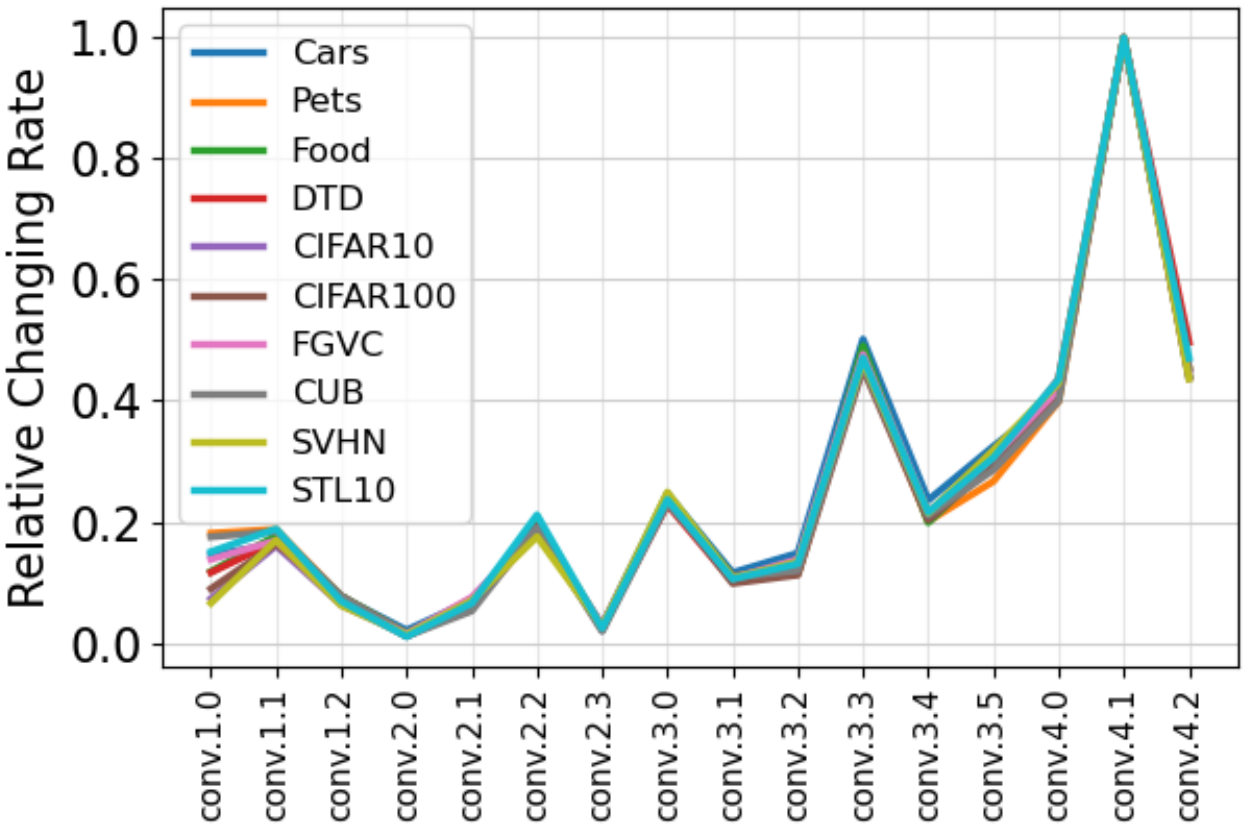}}
\subfigure[Resnet101]{
\includegraphics[width=0.32\textwidth]{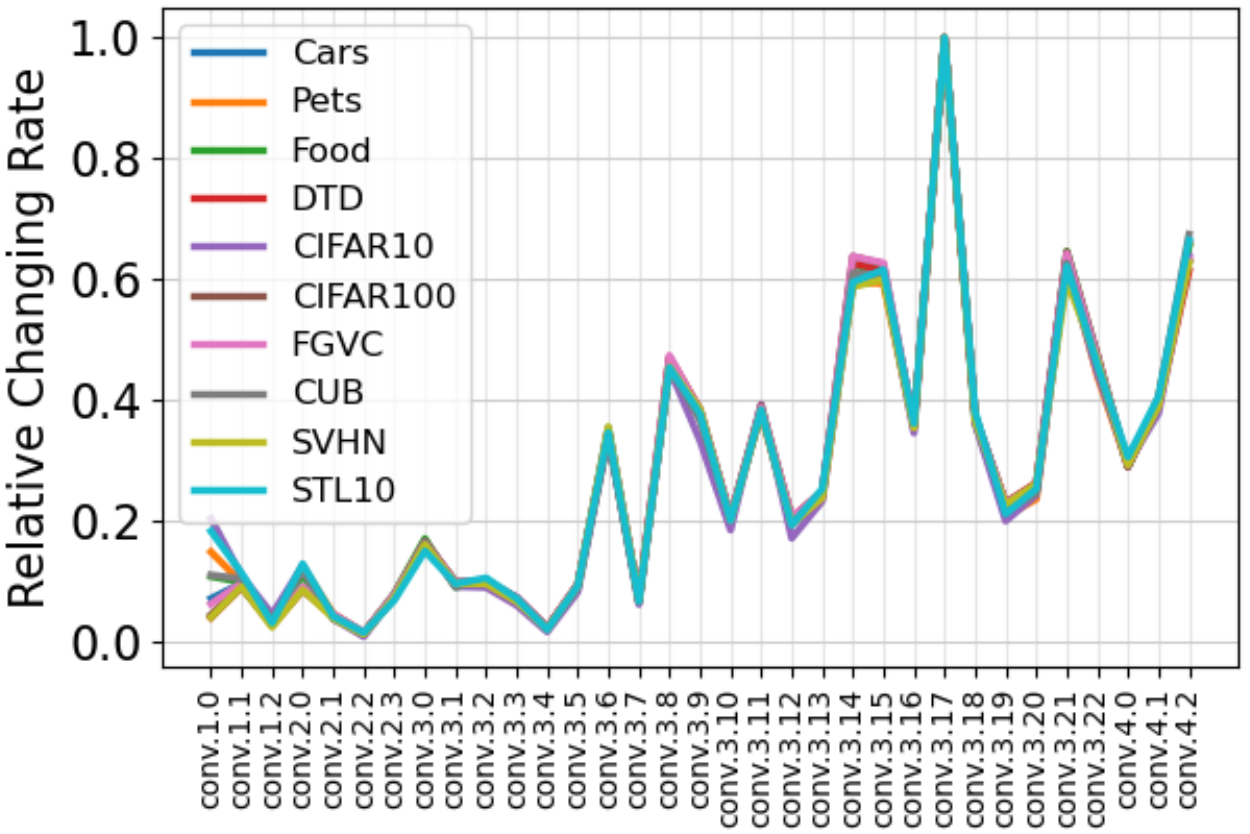}}
\subfigure[ViT16]{
\includegraphics[width=0.32\textwidth]{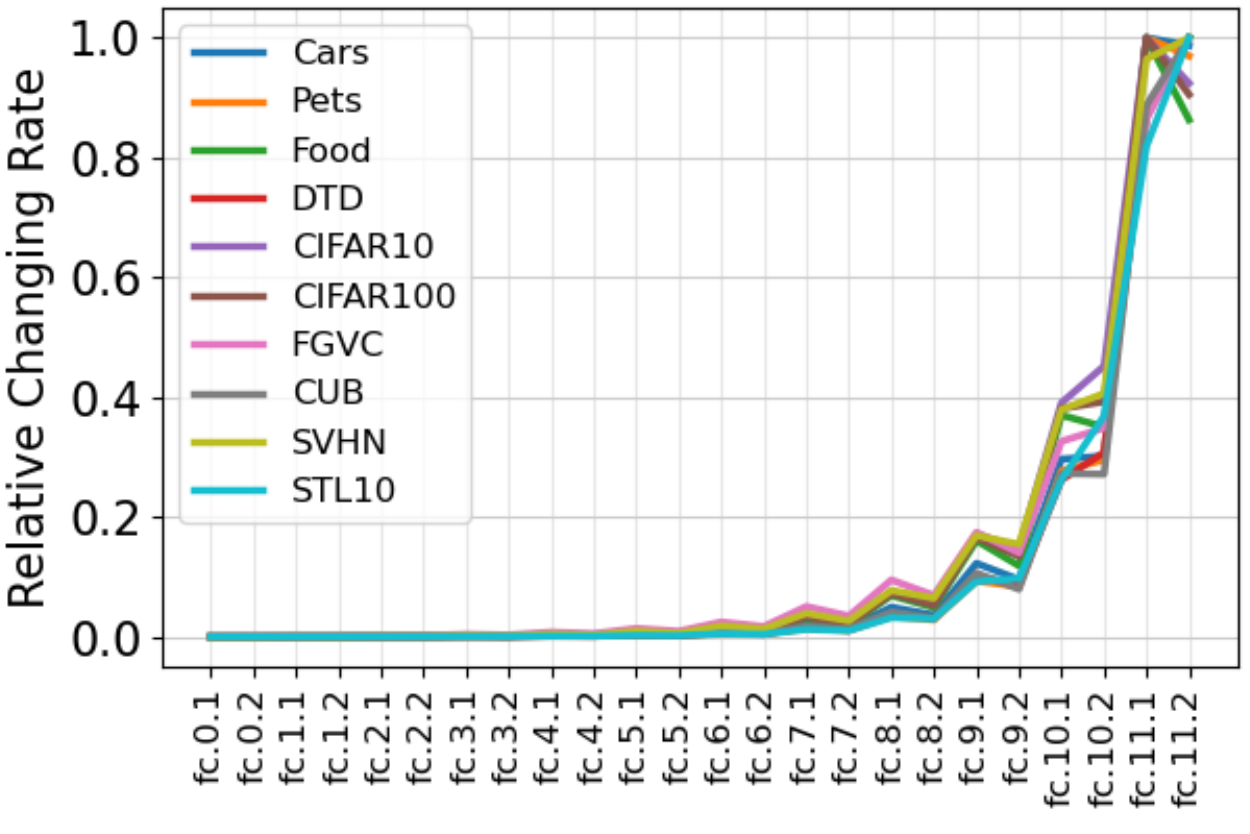}}
\caption{The ordinate represents the Frobenius norm of the difference between the parameters of the fine-tuned model and its corresponding pre-trained model, which is scaled into a range from 0 to 1 for easy comparison. The abscissa represents the level of the layer. Note that Resnet50 and Resnet101~\cite{he2016resnet} are pre-trained by SimCLRv2~\cite{chen2020simple}, and ViT16~\cite{vaswani2017transformer} is pre-trained by MAE~\cite{he2021masked}.} 
\label{fig:parameter}
\end{figure}

\textbf{Low-Level Layer Lifting Attack (L4A).}
Our method is motivated by the findings in Fig.~\ref{fig:parameter} that the higher the level of the layers, the more their parameters change during fine-tuning. This is also consistent with the knowledge that the low-level convolutional layer acts as an edge detector that extracts low-level features like edges and textures and has little high-level semantic information~\cite{oquab2014transfer,yosinski2014transferable}. Since images from different datasets share the same low-level features, the parameters of these layers can be preserved during fine-tuning. In contrast, the attack algorithms based on the high-level layers or the scores predicted by the model may not transfer well in such a cross-finetuning setting, as the feature spaces of high-level layers are easily distorted during fine-tuning. The basic method of L4A can be formulated as the following problem:
\begin{equation}\label{eq:L4Abase}
\min_{\vect{\delta}} L_{base}(f_{\vect{\theta}},\vect{x},\vect{\delta}) = -\mathbb{E}_{\vect{x}\sim D_p}\bigg[\|f_{\vect{\theta}}^{k}(\vect{x}+\vect{\delta})\|_F^{2}\bigg],
\end{equation}
where $\|\cdot\|_F$ denotes the Frobenius norm of the input tensor.
In our experiments, we find the lower the layer, the better it performs, so we choose the first layer as default, such that $k = 1$.
As Eq.~\eqref{eq:L4Abase} is usually a sophisticated non-convex optimization problem, we solve it using stochastic gradient descent method.

We also find that fusing the adversarial loss of the consecutive low-level layers can boost the performance, which gives L4A\textsubscript{fuse} method as solving:
\begin{equation}
\min_{\vect{\delta}}  L_{fuse}(f_{\vect{\theta}},\vect{x},\vect{\delta}) = -\mathbb{E}_{\vect{x}\sim D_p}\bigg[\|f_{\vect{\theta}}^{k_1}(\vect{x}+\vect{\delta})\|_F^{2}+\lambda\cdot\|f_{\vect{\theta}}^{k_2}(\vect{x}+\vect{\delta})\|_F^{2}\bigg],
\end{equation}
where $f_{\vect{\theta}}^{k_1}(\vect{x}+\vect{\delta})$ and $f_{\vect{\theta}}^{k_2}(\vect{x}+\vect{\delta})$ refers to the $k_1$-th and $k_2$-th layers' feature maps of $f_{\vect{\theta}}$ respectively, $\lambda$ is a balancing hyperparameter. We set $k_1 = 1$ and $k_2 = 2$ as default.

\begin{wrapfigure}{r}{0.33\textwidth}
\vspace{-2ex}
\centering
\includegraphics[width=0.97\linewidth]{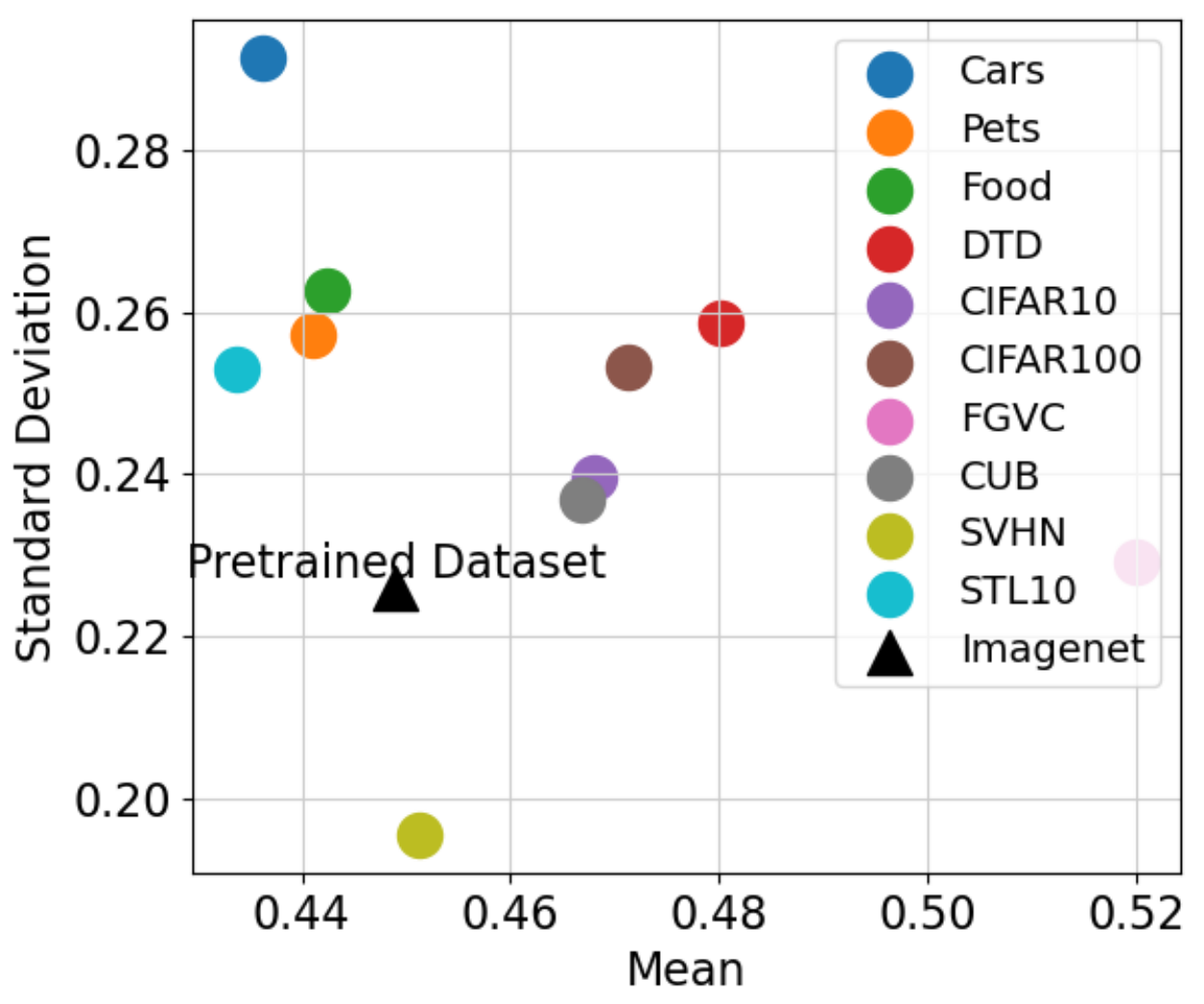}
\vspace{-2ex}
\caption{Datasets' statistics.}
\label{fig:mean_std}
\end{wrapfigure}
\textbf{Uniform Gaussian Sampling.}
Nowadays, most state-of-the-art networks apply batch normalization~\cite{ioffe2015batch} to input images for better performance. Thus, the datasets' statistics become an essential factor for training. As shown in Fig.~\ref{fig:mean_std}, the distribution of the downstream datasets can vary significantly compared to that of the pre-training dataset. However, traditional data augmentation techniques~\cite{yun2019cutmix,inoue2018data} are limited to the pre-training domain and cannot alleviate the problem. Thus, we propose sampling Gaussian noises with various means and deviations to avoid overfitting.
Combining the base loss using the pre-training dataset and the new loss using uniform Gaussian noises gives the L4A\textsubscript{ugs} method as follows:
\begin{equation}
\label{eq:L4Augs}
\min_{\vect{\delta}}  L_{ugs}(f_{\vect{\theta}},\vect{x},\vect{\delta}) =
-\mathbb{E}_{\vect{\vect{\mu}},\vect{\sigma},\vect{n}_0\sim N\left(\vect{\mu},\vect{\sigma}\right) }\bigg\{\mathbb{E}_{\vect{x}\sim \mathcal{D}_p}
\bigg[{\|f_{\vect{\theta}}^{k}(\vect{x}+\vect{\delta})\|_F^{2}}+\lambda\cdot{\|f_{\vect{\theta}}^{k}(\vect{n}_0+\vect{\delta})\|_F^{2}}\bigg]\bigg\},
\end{equation}
where $\vect{\vect{\mu}}$ and $\vect{\vect{\sigma}}$ are drawn from the uniform distribution $U\left(\vect{\mu}_l,\vect{\mu}_h\right)$ and $U\left(\vect{\sigma}_l,\vect{\sigma}_h\right)$, respectively, and $\vect{\mu}_l$,  $\vect{\mu}_h$, $\vect{\sigma}_l$, $\vect{\sigma}_r$ are four hyperparameters. 


\section{Experiments}\label{sec:experiment}

We provide some experimental results in this section. More results can be found in Appendix. Our code is publicly available at \url{https://github.com/banyuanhao/PAP}.

\subsection{Settings}

\textbf{Pre-training methods.} SimCLR~\cite{chen2020simple,chen2020big} uses the Resnet~\cite{he2016resnet} backbone and pre-trains the model by contrastive learning. We download pre-trained parameters of Resnet50 and Resnet101\footnote{\url{https://github.com/google-research/simclr}\label{foot1}} to evaluate the generalization ability of our algorithm on different architectures. We also adopt MOCO \cite{he2020moco} with the backbone of Resnet50\footnote{\url{https://dl.fbaipublicfiles.com/moco/}}. Besides convolutional neural networks, transformers~\cite{vaswani2017transformer} attract much attention nowadays for their competitive performance. Based on transformers and masked image modeling, MAE~\cite{he2021masked} becomes a good alternative for pre-training. We adopt the pre-trained ViT-base-16 model\footnote{\url{https://github.com/facebookresearch/mae}\label{foot2}}. Moreover, vision-language pre-trained models are gaining popularity these days. Thus we also choose CLIP~\cite{radford2021clip}\footnote{\url{https://github.com/openai/CLIP}} for our study. 
We report the results of SimCLR and MAE in Section~\ref{mainresults}. More results on CLIP and MOCO can be found in Appendix~\ref{appaddpre}.

\begin{table}[!t]\footnotesize
\setlength{\tabcolsep}{5pt}
\caption{The attack success rate (\%) of various attack methods against  \textbf{Resnet101} pre-trained by \textbf{SimCLRv2}. Note that C10 stands for CIFAR10 and C100 stands for CIFAR100.}
\begin{tabular}{l||ccccccc|ccc|c}
\hline
ASR
& Cars                                    & Pets                                    & Food                           & DTD                                     & FGVC                                    & CUB                                     & SVHN                                   & C10                                 & C100                                & STL10                                   & AVG                                     \\ \hline\hline
FFF\textsubscript{no}                           & 43.81                                 & 38.62                                 & 49.95                        & 63.24                                 & 85.57                                 & 48.38                                 & 12.55                                 & 8.53                                  & 77.74                                 & 57.11                                 & 48.55                                 \\
FFFF\textsubscript{mean}                           & 33.93                                 & 31.37                                 & 41.77                        & 52.66                                 & 78.94                                 & 45.00                                 & \textbf{14.85}                       & 14.42                                 & 72.59                                 & 56.66                                 & 44.22                                 \\
FFF\textsubscript{one}                            & 31.87                                 & 29.74                                 & 39.25                        & 46.92                                 & 74.17                                 & 43.87                                 & 9.24                                  & 11.77                                 & 65.61                                 & 50.21                                 & 40.26                                 \\
DR                            & 36.28                                 & 35.54                                 & 47.43                        & 47.45                                 & 75.00                                 & 44.15                                 & 12.05                                 & 21.35                                 & 65.39                                 & 41.65                                 & 42.63                                 \\
SSP                            & 32.89                                 & 30.50                                 & 43.12                        & 45.85                                 & 82.57                                 & 45.55                                 & 8.69                                  & 11.66                                 & 65.80                                 & 40.91                                 & 40.75                                 \\
ASV                            & 60.75                                 & 19.84                                 & 36.33                        & 56.22                                 & 84.16                                 & 55.82                                 & 7.11                                  & 7.29                                  & 58.10                                 & 80.89                                 & 46.64                                 \\
UAP                            & 48.70                                 & 36.55                                 & 60.80                        & 63.40                                 & 76.06                                 & 52.64                                 & 8.46                                  & 8.53                                  & 52.35                                 & 31.15                                 & 43.86                                 \\
{\color[HTML]{3531FF} UAPEPGD} & {\color[HTML]{3531FF} 94.12}          & {\color[HTML]{3531FF} 66.66}          & {\color[HTML]{3531FF} 61.30} & {\color[HTML]{3531FF} 72.55}          & {\color[HTML]{3531FF} 70.34}          & {\color[HTML]{3531FF} 82.72}          & {\color[HTML]{3531FF} 13.88}          & {\color[HTML]{3531FF} \textbf{61.65}} & {\color[HTML]{3531FF} 20.04}          & {\color[HTML]{3531FF} 50.13}          & {\color[HTML]{3531FF} 59.34}          \\ \hline
L4A\textsubscript{base}                       & 94.07                                 & 61.57                                 & 71.23                        & 69.20                                 & \textbf{96.28}                                 & 81.07                                 & 11.70                                 & 12.68                                 & 80.57                                 & 90.49                                 & 66.89                                 \\
L4A\textsubscript{fuse}                       & 90.98                                 & 88.53                                 & \textbf{80.65}               & 74.31                                 & 93.79                                 & 91.23                                 & 11.40                                 & 17.40                                 & \textbf{80.98}                                 & 89.69                                 & 67.10                                 \\
{\color[HTML]{FE0000} L4A\textsubscript{ugs}} & {\color[HTML]{FE0000} \textbf{94.24}} & {\color[HTML]{FE0000} \textbf{94.99}} & {\color[HTML]{FE0000} 78.28} & {\color[HTML]{FE0000} \textbf{77.23}} & {\color[HTML]{FE0000} 92.92} & {\color[HTML]{FE0000} \textbf{91.77}} & {\color[HTML]{FE0000} \textbf{11.40}} & {\color[HTML]{FE0000} 14.60}          & {\color[HTML]{FE0000} 76.50} & {\color[HTML]{FE0000} \textbf{90.05}} & {\color[HTML]{FE0000} \textbf{72.20}} \\ \hline
\end{tabular}\label{tab:res101}
\centering
\end{table}

\textbf{Datasets and Pre-processing.} We adopt the ILSVRC 2012  dataset~\cite{russakovsky2015imagenet} to generate PAPs, which are also used to pre-train the models. We mainly evaluate PAPs on image classification tasks, which are the same as the settings of SimCLRv2. Ten fine-grained and coarse-grained datasets are used to test the cross-finetuning transferability of the generated PAPs. We load these datasets from torchvision (Details in Appendix~\ref{appdataset}). Before feeding the images to the model, we resize them to $256\times256$ and then center crop them into $224\times224$.

\textbf{Compared methods.} We choose UAP~\cite{moosavi2017universal} to test whether image-agnostic attacks also bear good cross-finetuning transferability. Since UAP needs final classification predictions of the inputs, we fit a linear head on the pre-trained feature extractor. Furthermore, by integrating the moment term into the iterative method, UAPEPGD~\cite{deng2020uapepgd} is believed to enhance cross-model transferability. Thus, we adopt UAPEPGD to study the connection between cross-model and cross-finetuning transferability. As our algorithm is based on the feature level, other feature attacks (including FFF~\cite{mopuri2017fff}, ASV~\cite{khrulkov2018asv}, DR~\cite{lu2020DR}, SSP~\cite{naseer2020ssp}) are chosen for comparison.

\textbf{Default settings and Metric.}
Unless otherwise specified, we choose a batch size of $16$ and a step size of $0.0002$. All the perturbations should be within the bound of $0.05$ under the $\ell_{\infty}$ norm. We evaluate the perturbations at the iterations of $1,000$, $5,000$, $30,000$, and $60,000$, and report the best performance. We show the results in the measure of attack success rates (ASR), representing the classification error rate on the whole testing dataset after adding the perturbations to the legitimate images. 

\subsection{Main results}\label{mainresults}

We craft pre-trained adversarial perturbations (PAPs) for three pre-trained models (i.e., Resnet50 by SimCLRv2, Resnet101 by SimCLRv2, ViT16 by MAE) and evaluate the attack success rates on ten downstream tasks. The results are shown in Table~\ref{tab:res101}, Table~\ref{tab:res50}, and Table~\ref{tab:vit}, respectively. Note that the first seven datasets are fine-grained, while the last three are coarse-grained ones. We mark the best results for each dataset in \textbf{bold}, and the best baseline in \textcolor{blue}{blue}. We highlight the results of L4A\textsubscript{ugs} in \textcolor{red}{red} to emphasize that the L4A attack equipped with Uniform Gaussian Sampling shows great \emph{cross-finetuning transferability} and performs best.

\begin{table}[t]\footnotesize
\setlength{\tabcolsep}{5pt}
\caption{The attack success rate (\%) of various attack methods against \textbf{Resnet50} pre-trained by \textbf{SimCLRv2}. Note that C10 stands for CIFAR10, and C100 stands for CIFAR100.}
\centering
\begin{tabular}{l||ccccccc|ccc|c}
\hline
ASR                        & Cars                                    & Pets                                    & Food                                   & DTD                                     & FGVC & CUB                                     & SVHN                              & C10                                 & C100                                & STL10                                   & AVG                                     \\ \hline\hline
FFF\textsubscript{no}                        & 26.91                                 & 30.83                                 & 43.28                                 & 48.99   & 41.30                                                 &38.23                                  & 79.00                                      & 68.50                                 & 44.67                                 & 16.95                                 & 43.86                                 \\
FFF\textsubscript{mean}                      & 36.75                                 & 33.88                                 & 45.26                                 & 50.15   & 53.13                                   &77.22                                 & 52.02                                                    & 82.41                                 & 68.10                                 & 22.11                                 & 52.10                                 \\
FFF\textsubscript{one}                      & 37.88                                 & 35.30                                 & 52.79                                 & 59.52     & 59.62                                            &57.04                                 &80.33                                          & 75.40                                 & 53.58                                 & 18.31                                 & 52.98                                 \\
DR                             & 38.64                                 & 34.42                                 & 50.04                                 & 45.53                                      & 39.80         &75.67                                 &47.88                                           & 76.05                                 & 60.57                                 & 13.98                                 & 48.26                                 \\
SSP                            & 41.70                                 & 43.94                                 & 50.83                                 & 48.78                                 & 47.67                    &82.39                                 &48.38                                     & 86.95                                 & 66.23                                 & 19.56                                 & 53.64                                 \\
ASV                            & 74.47                                 & 36.93                                 & 45.85                                 & 73.51                               & 64.89                        &92.29                                 &73.16                                   & 45.14                                 & 53.60                                 & 22.02                                 & 58.19                                 \\
UAP                            & 44.86                                 & 46.47                                 & 64.67                                 & 65.53                                & 49.63                       &82.32                                 &52.00                                   & 79.63                                 & 46.46                                 & 19.99                                 & 55.16                                 \\
{\color[HTML]{3531FF} UAPEPGD} & {\color[HTML]{3531FF} 66.29}          & {\color[HTML]{3531FF} 66.58}          & {\color[HTML]{3531FF} \textbf{81.11}} & {\color[HTML]{3531FF} 69.52}          &{\color[HTML]{3531FF} 87.91}          & {\color[HTML]{3531FF} 59.07}  &{\color[HTML]{3531FF} 69.16}         & {\color[HTML]{3531FF} \textbf{87.84}} & {\color[HTML]{3531FF} 68.26}          & {\color[HTML]{3531FF} 37.12}          & {\color[HTML]{3531FF} 69.28}          \\ \hline
LLLL\textsubscript{base}                           & 94.86                                 & 56.30                                 & 61.31                                 & 75.37           & 67.61                                     &94.87                                 &81.45                                          & 68.25                                 & 77.04                                 & 34.56                                 & 66.89                                 \\
L4A\textsubscript{fuse}                           & 96.00                                 & 59.80                                 & 65.00                                 & 77.93                                          & 69.39    &\textbf{95.02}                                 &85.05                                  & 64.41                                 & 76.29                                 & 37.54                                 & 72.64                                 \\
{\color[HTML]{FE0000} L4A\textsubscript{ugs}} & {\color[HTML]{FE0000} \textbf{96.13}} & {\color[HTML]{FE0000} \textbf{79.15}} & {\color[HTML]{FE0000} 74.87}          & {\color[HTML]{FE0000} \textbf{82.18}}  & {\color[HTML]{FE0000} \textbf{78.73}} &{\color[HTML]{FE0000} 94.45} &{\color[HTML]{FE0000} \textbf{95.29}} & {\color[HTML]{FE0000} 55.03}          & {\color[HTML]{FE0000} \textbf{77.10}} & {\color[HTML]{FE0000} \textbf{45.09}} & {\color[HTML]{FE0000} \textbf{77.80}} \\ \hline
\end{tabular}\label{tab:res50}
\end{table}

\begin{table}[t]\footnotesize
\setlength{\tabcolsep}{5pt}
\caption{ The attack success rate (\%) of various attack methods against \textbf{ViT16} pre-trained by \textbf{MAE}. Note that C10 stands for CIFAR10 and C100 stands for CIFAR100.}
\centering
\begin{tabular}{l||ccccccc|ccc|c}
\hline
ASR
& Cars                                    & Pets                                    & Food                           & DTD                                     & FGVC                                    & CUB                                     & SVHN                                    & C10                                 & C100                                & STL10                                   & AVG                                     \\ \hline\hline
{\color[HTML]{3531FF} FFF\textsubscript{no}} & {\color[HTML]{3531FF} 64.31}          & {\color[HTML]{3531FF} 88.21}          & {\color[HTML]{3531FF} 95.04}          & {\color[HTML]{3531FF} 88.18} & {\color[HTML]{3531FF} 81.91}          & {\color[HTML]{3531FF} 92.94} & {\color[HTML]{3531FF} 76.10} & {\color[HTML]{3531FF} 49.48}          & {\color[HTML]{3531FF} 79.83}          & {\color[HTML]{3531FF} 60.91}          & {\color[HTML]{3531FF} 77.69}          \\
FFF\textsubscript{mean}                      & 40.39                                  & 67.54                                  & 54.10                                  & 61.38                         & 71.47                                  & 73.39                                  & \textbf{92.96 }                        & 86.88                                  & 94.55                                  & 67.90                              & 71.06                                  \\
FFF\textsubscript{one}                      & 48.36                                  & 77.89                                  & 60.06                                  & 64.04                         & 75.67                                  & 74.09                                  & 92.33                        & 86.13                                  & 94.48                                  & 70.40                                  & 74.35                                  \\
DR                             & 37.02                                  & 23.84                                  & 59.54                                  & 44.73                         & 28.01                                  & 10.41                                  & 14.30                         & 16.66                                  & 14.12                                  & 21.54                                  & 27.02                                  \\
SSP                            & 44.15                                  & 73.31                                  & 85.42                                  & 72.82                         & 52.57                                  & 63.10                                  & 52.45                        & 27.94                                  & 25.32                                  & 36.90                                  & 53.40                                  \\
ASV                            & 38.46                                  & 10.17                                  & 37.49                                  & 48.31                         & 29.14                                  & 4.97                                   & 8.41                          & 17.04                                  & 11.34                                  & 21.14                                  & 22.64                                  \\
UAP                            & 62.71                                  & 58.90                                  & 89.90                                  & 74.92                         & 44.69                                  & 39.56                                  & 47.65                         & 47.77                                  & 33.80                                  & 51.70                                 & 55.16                                 \\
{\color[HTML]{000000} UAPEPGD} & {\color[HTML]{000000} 63.67}          & {\color[HTML]{000000} 73.09}          & {\color[HTML]{000000} 96.22}          & {\color[HTML]{000000} 76.69} & {\color[HTML]{000000} 57.78}          & {\color[HTML]{000000} 73.37}          & {\color[HTML]{000000} 79.84} & {\color[HTML]{000000} 45.89}          & {\color[HTML]{000000} 47.21}          & {\color[HTML]{000000} 55.79}          & {\color[HTML]{000000} 66.95}          \\ \hline
L4A\textsubscript{base}                       & 87.66                                 & 89.98                                 & 98.96                                 & \textbf{99.10}               & 99.33                                 & 84.06                                 & 86.99                        & 98.62                                 & 97.08                                 & 98.25                                 & 94.00                                 \\
L4A\textsubscript{fuse}                       & 83.24                                 & 89.57                                 & 98.87                                 & 98.77                        & 98.36                                 & \textbf{93.60}                                 & 89.85               & 98.64                                 & 95.72                                 & 97.53                                 & 94.42                                 \\
{\color[HTML]{FE0000} L4A\textsubscript{ugs}} & {\color[HTML]{FE0000} \textbf{96.49}} & {\color[HTML]{FE0000} \textbf{90.00}} & {\color[HTML]{FE0000} \textbf{98.97}} & {\color[HTML]{FE0000} 98.89} & {\color[HTML]{FE0000} \textbf{99.48}} & {\color[HTML]{FE0000} 84.01}          & {\color[HTML]{FE0000} 89.56} & {\color[HTML]{FE0000} \textbf{99.43}} & {\color[HTML]{FE0000} \textbf{97.27}} & {\color[HTML]{FE0000} \textbf{98.96}} & {\color[HTML]{FE0000} \textbf{95.30}} \\ \hline
                  
\end{tabular}\label{tab:vit}
\end{table}

A quick glimpse shows that our proposed methods outperform all the baselines by a large margin. For example, as can be seen from Table~\ref{tab:res101}, if the target model is Resnet101 pre-trained by SimCLRv2, the best competitor FFF\textsubscript{mean} achieves an average attack success rate of 59.34\%, while the villain L4A\textsubscript{base} can lift it up to \textbf{66.89\%} and the UGS technique further boosts the performance up to \textbf{72.20\%}. Moreover, the STL10 dataset is the hardest for PAPs to transfer among these tasks. However, L4A\textsubscript{ugs} can significantly improve the cross-finetuning transferability, achieving an attack success rate of 90.05\% and 98.96\% for Resnet101 and ViT16 in STL10, respectively. Another intriguing finding is that ViT16 with a transformer backbone shows severe vulnerabilities to PAPs. Although performing best on legitimate samples, they bear an attack success rate of \textbf{95.30\%} under the L4A\textsubscript{ugs} attack, closing to random outputs. These results reveal the serious security problem of the pre-training to fine-tuning paradigm and demonstrate the effectiveness of our method in such a problem setting.

\subsection{Ablation studies}

\subsubsection{Effect of the attacking layer}

We analyze the influence of attacking \emph{different intermediate layers} of the networks on the performance of our proposed L4A\textsubscript{base} in the pre-training domain (ImageNet) and the fine-tuning domains (ten downstream tasks). To this end, we divide Resnet50, Resnet101, and ViT into five blocks (Details in Appendix~\ref{modelarchi}) and conduct our algorithm on them. Note that for the fine-tuning domains, the average attack success rates on the ten datasets are reported.

\begin{figure}[t]
\centering  
\subfigure{
\includegraphics[width=0.47\textwidth]{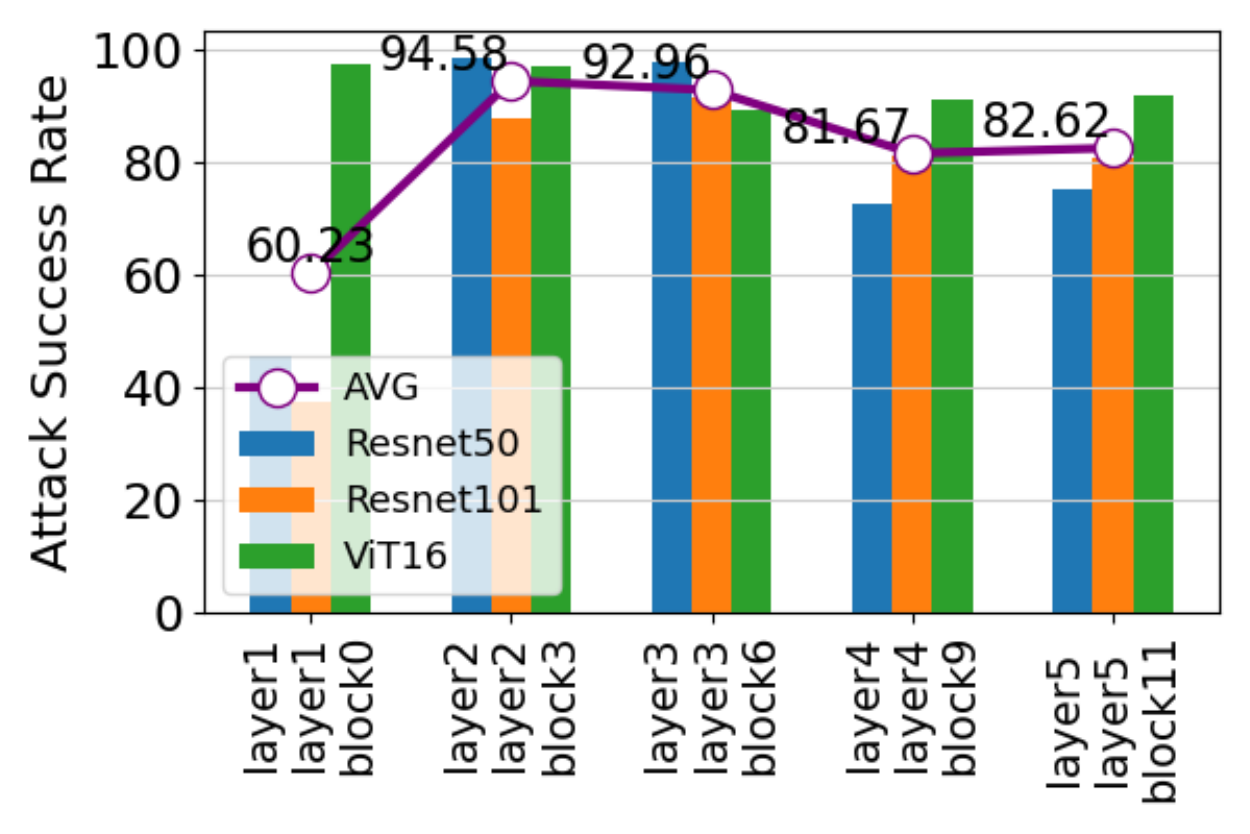}}
\subfigure{
\includegraphics[width=0.47\textwidth]{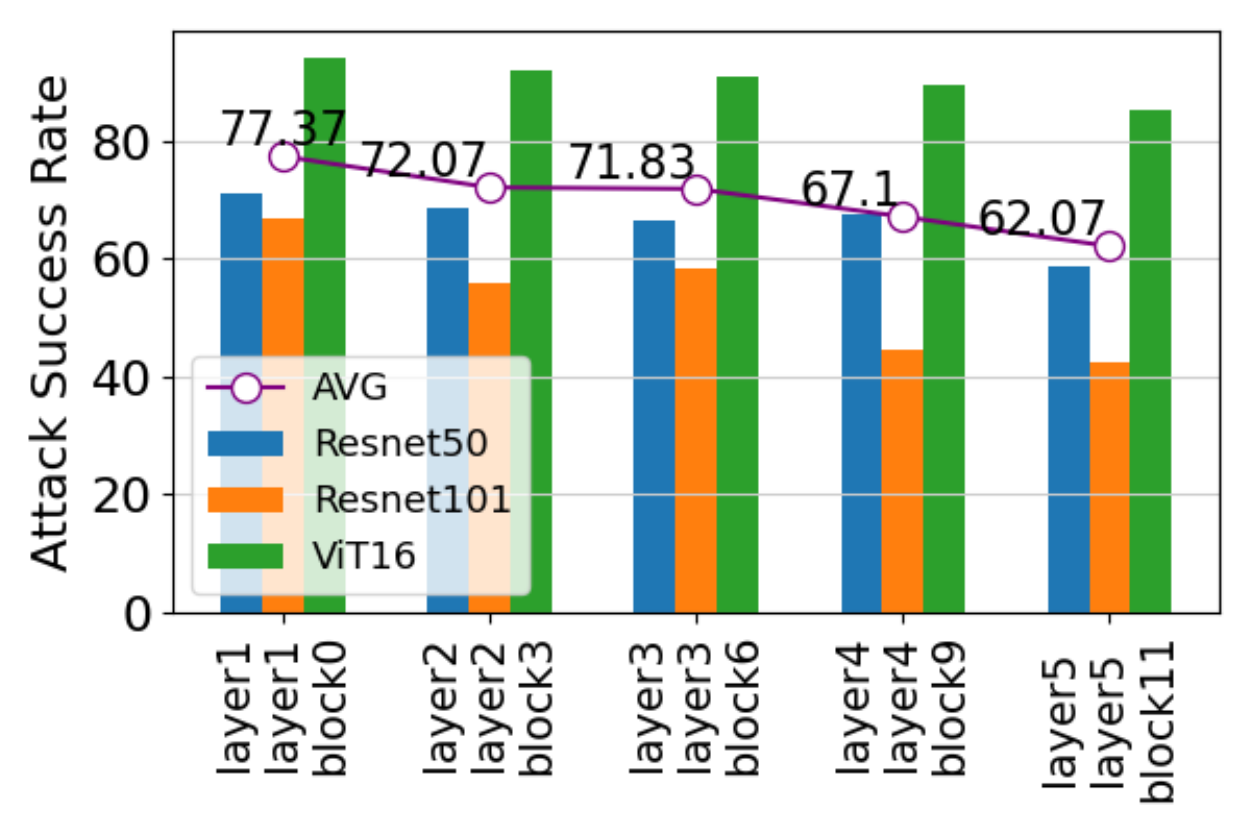}}
\caption{The attack success rates (\%) of L4A\textsubscript{base} when using different layers. We show the results on the pre-training domain (\textbf{Left}) and fine-tuning domains (\textbf{Right}).}
\label{fig:imageandtuned}
\end{figure}

As shown in Fig.~\ref{fig:imageandtuned}, the lower the level we choose to attack, the better our algorithm performs in the fine-tuning domains. Moreover, for the pre-training domain, attacking the middle layers of the networks results in a higher attack success rate compared to the top and bottom layers, which is also reported in existing works~\cite{lu2020std,naseer2020ssp,zhang2022beyond}.
These results reveal the intrinsic property of the pre-training to fine-tuning paradigm. As the lower-level layers change less during the fine-tuning procedure, attacking the low-level layer becomes more effective when generating adversarial perturbations in the pre-training domain rather than the middle-level layers.

\subsubsection{Effect of uniform Gaussian sampling}
\begin{figure}[t]
\centering  
\subfigure[Resnet101]{
\label{fig3.1}
\includegraphics[width=0.32\textwidth]{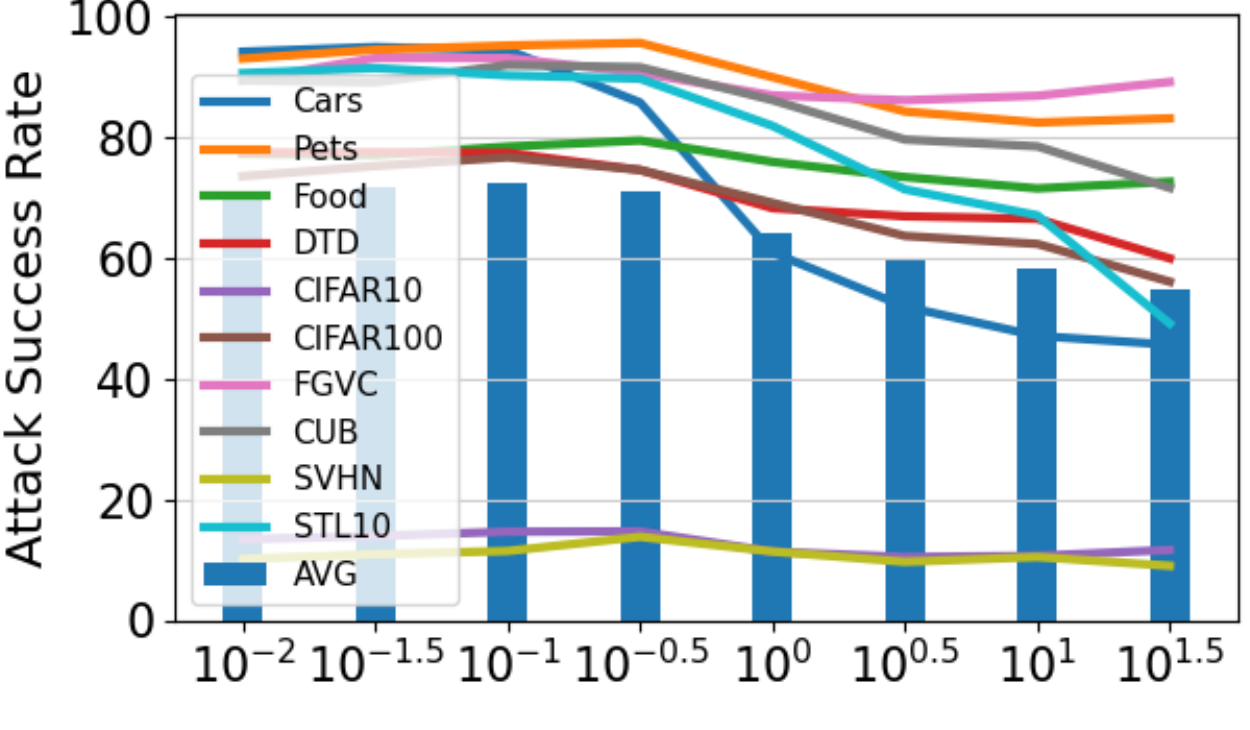}}
\subfigure[Resnet50]{
\label{fig3.2}
\includegraphics[width=0.32\textwidth]{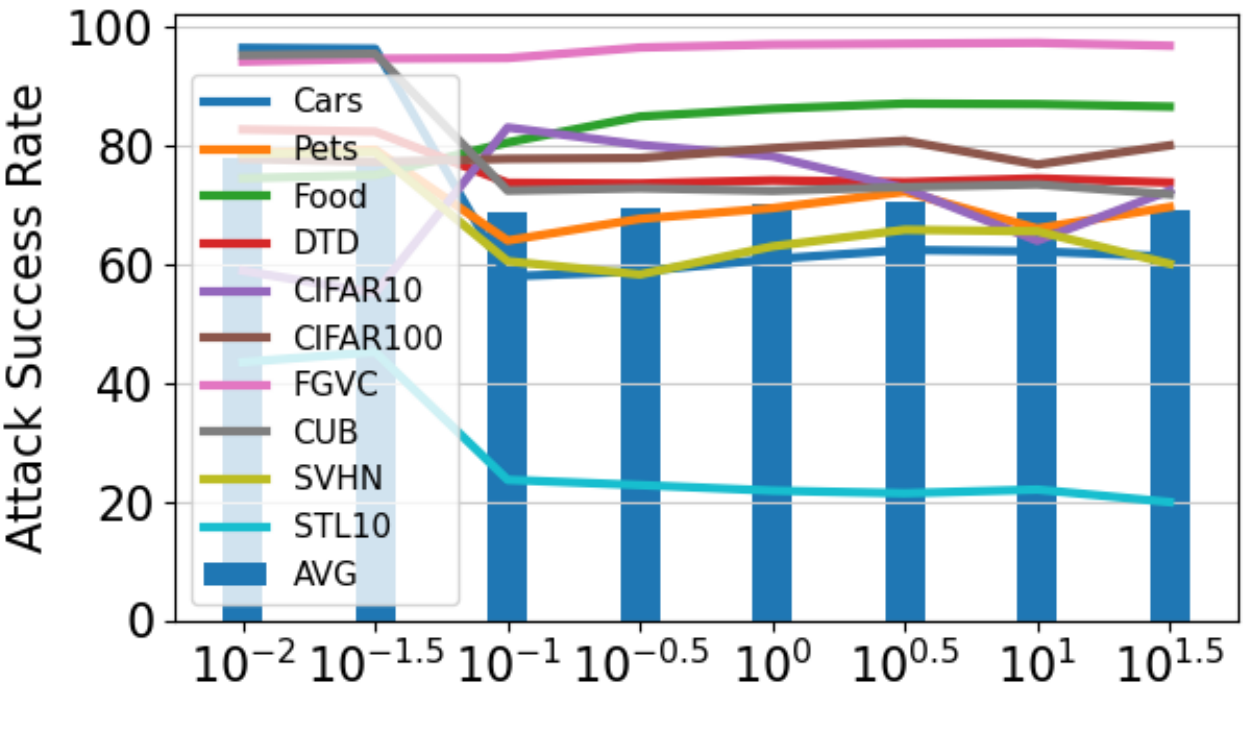}}
\subfigure[ViT16]{
\label{fig3.3}
\includegraphics[width=0.32\textwidth]{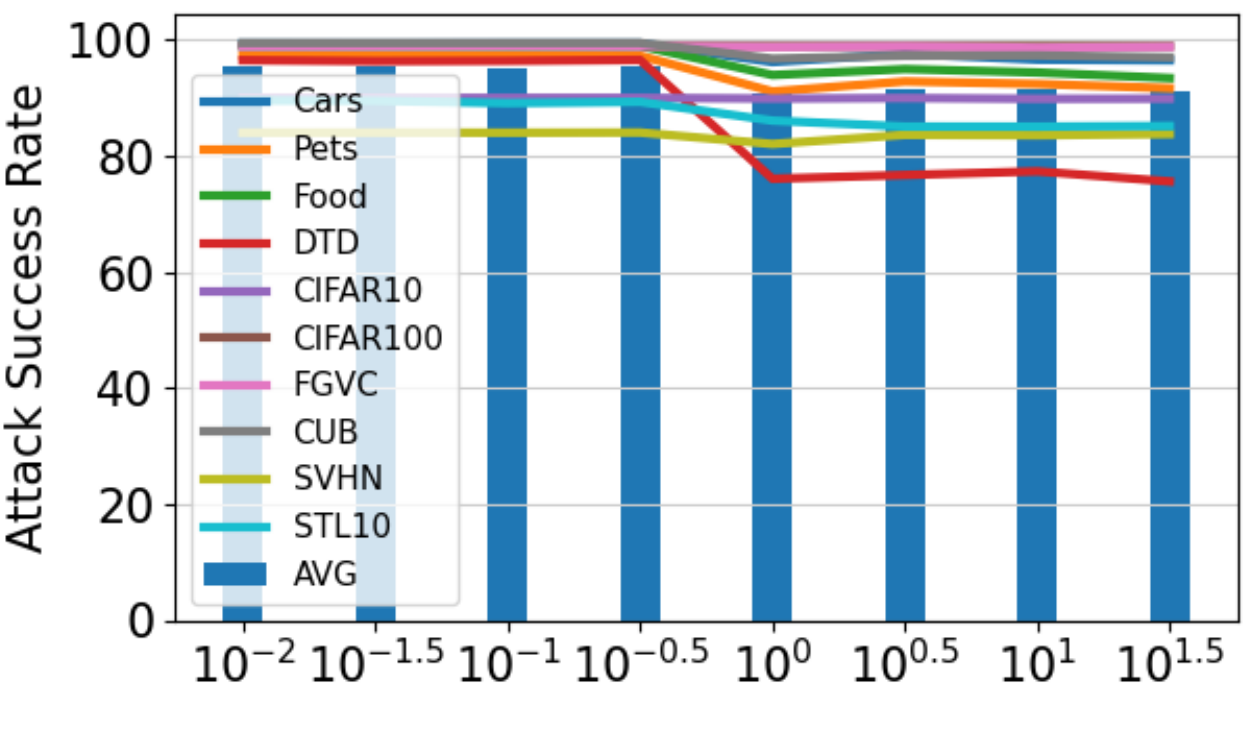}}
\caption{The attack success rate (\%) of different hyperparameter $\lambda$ in L4A\textsubscript{ugs} for different models.} 
\label{fig:lamuda_ugs}
\end{figure}

We set $\vect{\mu}_l,\vect{\mu}_h,\vect{\sigma}_l,\vect{\sigma}_h$ as 0.4, 0.6, 0.05, 0.1 for all the three models as they perform best. To discuss the effect of the hyperparameter $\lambda$ in Eq.~\eqref{eq:L4Augs} fusing the base loss and the UGS loss, we select the values with a grid of 8 logarithmically spaced learning rates between $10^{-2}$ and $10^{2}$. The results are shown in Fig.~\ref{fig:lamuda_ugs}. As 
shown in Fig.~\ref{fig:lamuda_ugs}\ref{fig3.1}, the best attack success rate is achieved when $\lambda=10^{-0.5}$ on Resnet101, boosting the performance by 1.52\% compared to only using the Gaussian noises.

\begin{wraptable}{r}{6cm}\footnotesize
\centering
\vspace{-17pt}
\caption{Fixed datasets' statistics}
\vspace{+5pt}
\begin{tabular}{l|lll}
\hline
Model    & R101    & R50     & MAE     \\ \hline
None    & 66.95 & 71.16 & 94.00 \\
ImageNet  & 68.39 & 69.50 & 95.30 \\
Uniform  & 72.20 & 77.80 & 95.30 \\ \hline
\end{tabular}
\label{tab:ugs}
\vspace{-10pt}
\end{wraptable}

Furthermore, we study whether adopting the fixed statistics of the pre-training dataset (i.e., the mean and standard deviation of ImageNet) can help. We report the attack success rates (\%) in the Table~\ref{tab:ugs}, where \textbf{None} refers to using no data augmentation, \textbf{ImageNet} adopts the mean and standard deviation of ImageNet and \textbf{Uniform} samples a pair of mean and standard deviation from the uniform distribution. As can be seen from the table, \textbf{Uniform} outperforms \textbf{None} by 4.39\% on average, while \textbf{ImageNet} does not help, which means that our proposed UGS helps to avoid overfitting the pre-training domain.

\subsection{Visualization of feature maps}

\begin{wrapfigure}{h}{0.40\textwidth}
\vspace{-12pt}
\centering
\includegraphics[width=0.99\linewidth]{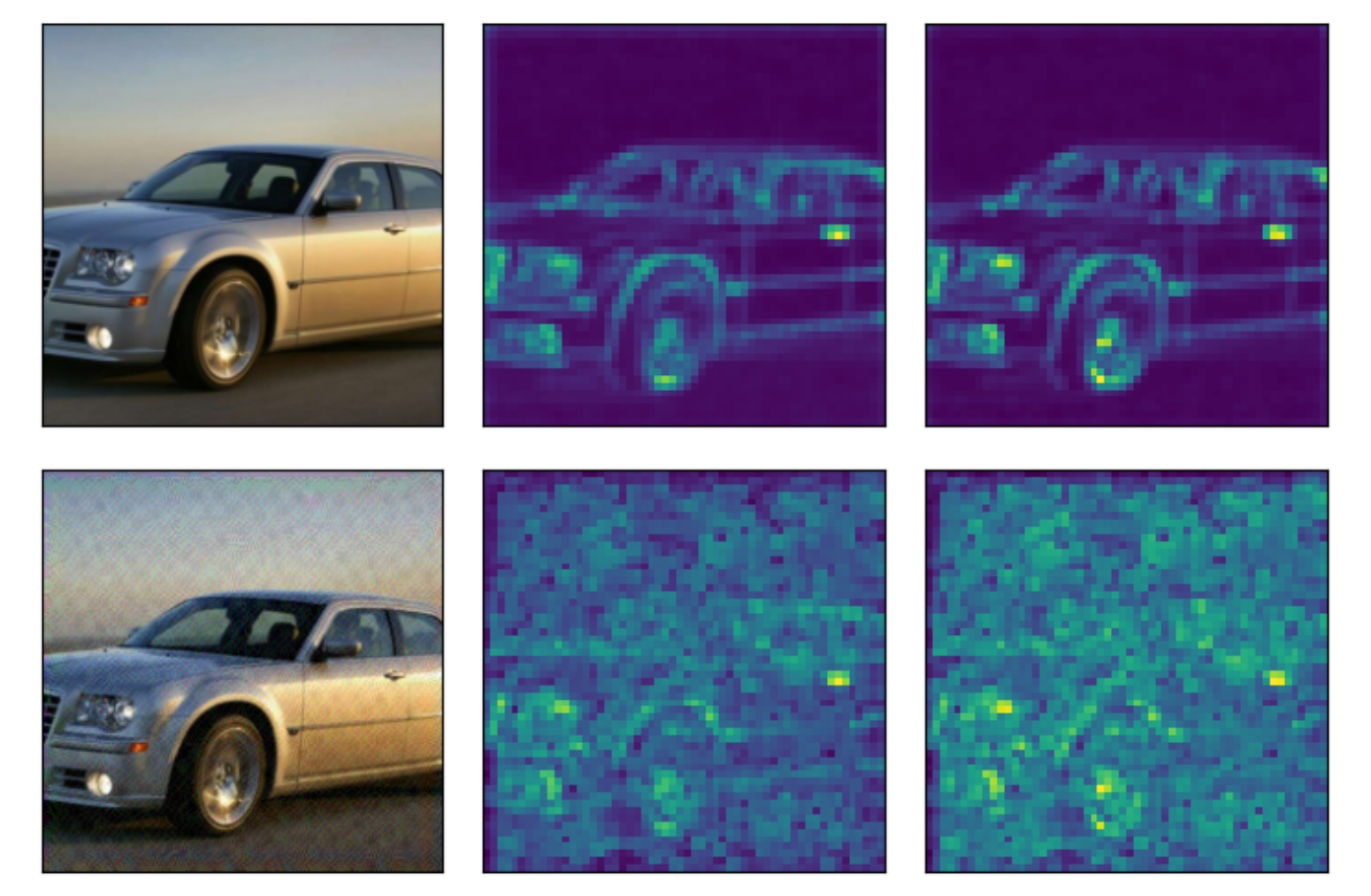}
\vspace{-16pt}
\caption{Visualization of feature maps.}
\label{fig6}
\vspace{-1pt}
\end{wrapfigure}
We show the feature maps before and after L4A\textsubscript{fuse} attack in Fig.~\ref{fig6}. The \textbf{Left} column shows the inputs of the model, while the  \textbf{Middle} and the \textbf{Right} show the feature map of the pre-trained model and the fine-tuned one, respectively. The \textbf{Upper} row represents the pipeline of a clean input in Cars, and the \textbf{Lower} shows that of adversarial ones. We can see from the upper row that fine-tuning the model can make it sensitive to the defining features related to the specific domain, such as tires and lamps. However, adding an adversarial perturbation to the image can significantly lift all the activations and finally mask the useful features. Moreover, the effect of our attack could be well preserved during fine-tuning and cheat the fine-tuned model into misclassification, stressing the safety problem of pre-trained models. 

\subsection{Trade-off between the clean accuracy and robustness}
\begin{wrapfigure}{r}{0.55\textwidth}
\vspace{-3ex}
\centering
\subfigure[Pets]{
\label{fig7.1}
\includegraphics[width=0.48\linewidth]{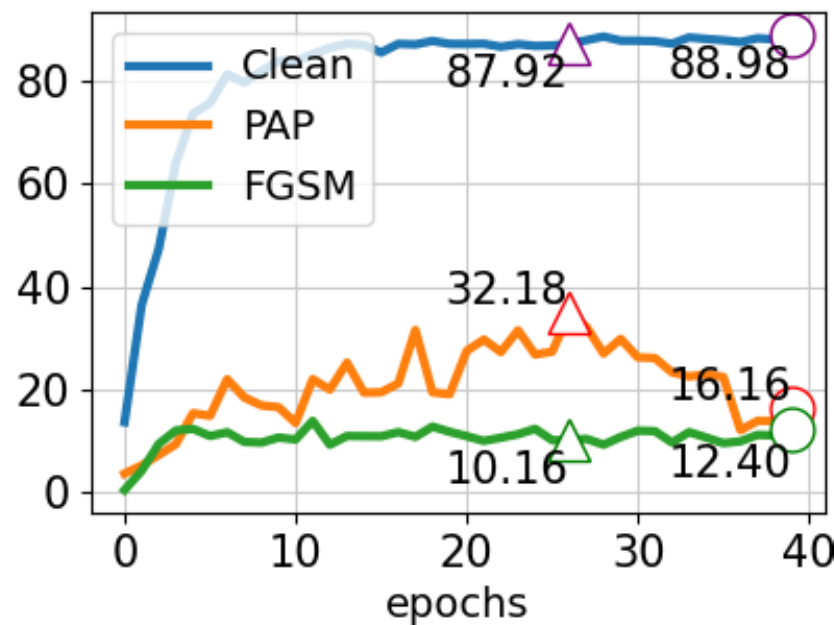}}
\subfigure[STL10]{
\label{fig7.2}
\includegraphics[width=0.48\linewidth]{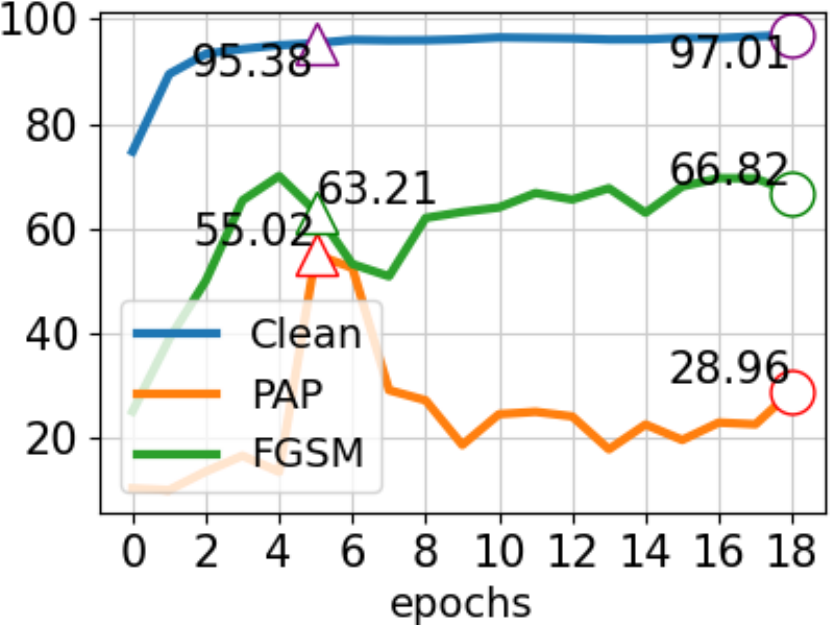}}
\vspace{-5pt}
\caption{Model accuracy (\%) on the Pets and STL10 datasets under clean inputs, PAP, and FGSM attack.}
\label{fig7}
\vspace{-3pt}
\end{wrapfigure}
We study the effect of fine-tuning epochs on the performance of our attack. To this end, we fine-tune the model until it reports the best result on the testing dataset, and then we plot the clean accuracy and the accuracy against FGSM and PAPs on Pets and STL10 in Fig.~\ref{fig7}.
The figure shows the clean accuracy and robustness of the fine-tuned model against PAPs are at odds. In Fig.~\ref{fig7}\ref{fig7.2}, the model shows the best robustness at epoch 5 in STL10, achieving an accuracy rate of 95.38\% and 55.02\% on clean and adversarial samples, respectively. However, the model does not converge until epoch 19. Though the process boosts the clean accuracy by \textbf{1.63\%}, it suffers a significant drop in robustness, as the accuracy on adversarial samples is lowered to \textbf{28.96\%}. Such findings reveal the safety problem of the pre-training to fine-tuning paradigm.

\section{Discussion}\label{sec:disscion}
In this section, we first introduce the gradient alignment and then use it to explain the effectiveness of our method. In particular, we show why our algorithms fall behind UAPs in the pre-training domain but have better cross-finetuning transferability when evaluated on the downstream tasks.

\subsection{Preliminaries}
\textbf{Gradient sequences.} Given a network $f_{{\vect{\theta}}_0}$ and a sample sequence $\{\vect{x}_1\, \vect{x}_2, \vect{x}_3, ...,\vect{x}_N\}_{\mathcal{D}}$ drawn from the dataset $\mathcal{D}$, let $\Delta\vect{\delta}_{\vect{\theta}_0,\mathcal{D}} = \{\Delta\vect{\delta}_{\vect{\theta}_0,\vect{x}_1},\Delta\vect{\delta}_{\vect{\theta}_0,\vect{x}_2},\Delta\vect{\delta}_{\vect{\theta}_0,\vect{x}_3},...,\Delta\vect{\delta}_{\vect{\theta}_0,\vect{x}_N}\}_{\mathcal{D}}$ be the sequence of gradients obtained when generating adversarial samples by the following equation:
\begin{equation}\label{eq:general}
 \Delta\vect{\delta}_{\vect{\theta}_0,{\vect{x}}_i} = \nabla_{\vect{\delta}}L(f_{\vect{\theta}_0},{\vect{x}}_{i},{\vect{\delta}}_i),\; \text{with } {\vect{\delta}}_i =  P_{\infty,\vect{\epsilon}}(\vect{\delta}_{i-1}+\Delta\vect{\delta}_{\vect{\theta}_0,{\vect{x}}_{i-1}}),
\end{equation}
where $L$ denotes the loss function of iterative attack methods like UAP, FFF, and L4A.

\begin{definition}[Gradient alignment] Given a dataset ${\mathcal{D}}$ and a model $f_{\vect{\theta}_0}$, the gradient alignment ${\mathcal{GA}}$ of an attack algorithm is defined as the expectation over the cosine similarity of $\Delta\vect{\delta}_{\vect{\theta}_0,\vect{x}_1}$ and $\Delta\vect{\delta}_{\vect{\theta}_0,\vect{x}_2}$, which can be formulated as Eq.~\eqref{eq:ga}
\begin{equation}\label{eq:ga}
 \mathcal{GA} = \mathbb{E}_{\vect{x}_1\sim \mathcal{D}, \vect{x}_2\sim \mathcal{D}}\left[\frac{\Delta\vect{\delta}_{\vect{\theta}_0,\vect{x}_1}\cdot{\Delta\vect{\delta}_{\vect{\theta}_0,\vect{x}_2}}}{\|\Delta\vect{\delta}_{\vect{\theta}_0,\vect{x}_1}\|_2\cdot\|{\Delta\vect{\delta}_{\vect{\theta}_0,\vect{x}_2}}\|_2}\right],
\end{equation}
where $\Delta\vect{\delta}_{\vect{\theta}_0,\vect{x}_1}$ and $\Delta\vect{\delta}_{\vect{\theta}_0,\vect{x}_2}$ are two consecutive elements obtained by Eq.~\eqref{eq:general}.
\end{definition}


Then \emph{the L4A algorithm bears a higher gradient alignment} (A strict definition and proof in a weaker form can be found in Appendix~\ref{app:proof}). In addition, we provide the results of a simulation experiment to justify it in Table~\ref{tab:ga}. We can see a negative correlation between the gradient alignment and the attack success rate on ImageNet. In contrast, a positive correlation exists between the gradient alignment and attack success rate in the fine-tuning domain.

\textbf{Effectiveness of the algorithm.} Given a pre-trained model $f_{\vect{\theta}}$ and the pre-training dataset $\mathcal{D}_p$, the generation of PAPs can be reformulated from Eq.~\eqref{optimization} to the following equation:
\begin{equation}
\label{reformulation}
 \max_{\vect{\delta}\in \; Span\{\Delta\vect{\delta}_{\vect{\theta},\mathcal{D}_p}\}}\mathbb{E}_{{\vect{x}}\sim \mathcal{D}_t}[F_{\vect{\theta}'}({\vect{x}})\neq F_{\vect{\theta}'}({\vect{x}}+\vect{\delta})], \; \text{s.t.}\;\|\vect{\delta}\|_{p}\leq \epsilon\;  \text{and }{\vect{x}}+\vect{\delta} \in \left [0,1\right],
\end{equation}
where $Span\{\Delta\vect{\delta}_{\vect{\theta},\mathcal{D}_p}\}$ denotes the subspace spanned by the elements of $\Delta\vect{\delta}_{\vect{\theta},\mathcal{D}_p}$. 
The optimal value $\vect{\delta}$ can be viewed as a linear combination of the elements in $\Delta\vect{\delta}_{\vect{\theta},\mathcal{D}_p}$, so it is the \emph{feasible region} of the equation. In particular, the feasible region of Eq.~\eqref{reformulation} is smaller than that of Eq.~\eqref{optimization}, which reflects that the stochastic gradient descent methods may not converge to the global maximum when it is not included in  $Span\{\Delta\vect{\delta}_{\vect{\theta},\mathcal{D}_p}\}$ and the local maximum of Eq.~\eqref{reformulation} in $Span\{\Delta\vect{\delta}_{\vect{\theta},\mathcal{D}_p}\}$ represents the effectiveness of the algorithm in the fine-tuning domain.

\makeatletter\def\@captype{figure}\makeatother
\begin{minipage}{0.52\textwidth}\
\centering
\includegraphics[width=1.0\linewidth]{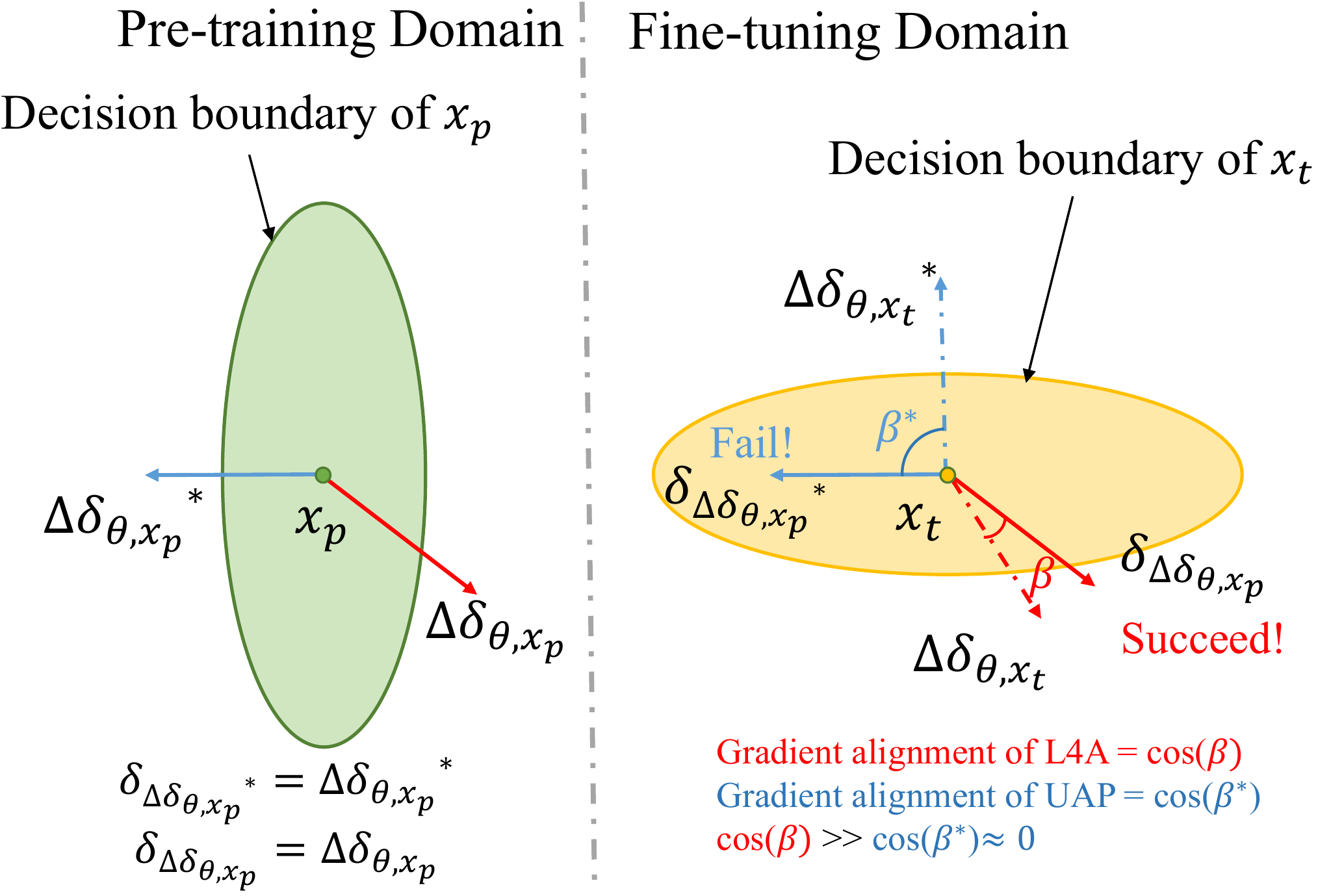}
\vspace{-15pt}
\caption{Illustration}\label{fig:ga}
\end{minipage}
\hspace{1ex}
\makeatletter\def\@captype{table}\makeatother
\begin{minipage}{0.45\textwidth}
\centering
\setlength{\tabcolsep}{1.5mm}
\begin{tabular}{l|ccc}
\hline
\textbf{Resnet101} & ${\mathcal{GA}}$              & ImageNet         & AVG              \\ \hline
FFF\textsubscript{no}                & 0.1221          & 38.64\%          & 48.55\%          \\
FFF\textsubscript{mean}                & 0.0194          & 48.37\%          & 44.22\%          \\
FFF\textsubscript{one}                & 0.1222          & 34.15\%          & 40.26\%          \\
DR                 & 0.0165          & 41.06\%          & 42.62\%          \\
UAP                & 0.0018          & 88.14\%          & 43.86\%          \\
UAPEPGD            & 0.0008          & \textbf{94.62\%} & 59.39\%          \\
SSP                & 0.0274          & 41.81\%          & 40.75\%          \\
L4A\textsubscript{base}                & \textbf{0.6125} & 37.44\%          & \textbf{66.95\%} \\ \hline
\end{tabular}
\caption{\textbf{Left:} Gradient alignments; \textbf{Middle:} Attack success rates on the pre-training dataset; \textbf{Right:} Average attack success rates on downstream tasks. See more details in Appendix~\ref{app:gradient}}\label{tab:ga}
\end{minipage}



\subsection{Explanation}
We aim to explain why the effectiveness of our algorithm is better in the fine-tuning domain and worse in the pre-training domain, as seen from Table~\ref{tab:ga}. 
Let $\Delta\vect{\delta}_{\vect{\theta},\mathcal{D}_p}$ and $\Delta\vect{\delta}_{\vect{\theta},\mathcal{D}_p}^{*}$ be the feasible zones of L4A and UAP obtained by feeding instances from $D_p$ into the pre-trained model $f_{\vect{\theta}}$, respectively. Similarly, we can define $\Delta\vect{\delta}_{\vect{\theta},\mathcal{D}_t}$ and  $\Delta\vect{\delta}_{\vect{\theta},\mathcal{D}_t}^{*}$. Meanwhile, denote $\vect{\delta}_{\Delta\vect{\delta}_{\vect{\theta},\mathcal{D}_p}}$ and $\vect{\delta}_{\Delta\vect{\delta}_{\vect{\theta},\mathcal{D}_p}}^{*}$ as the maxima in the pre-training domain obtained by L4A and UAP respectively. An illustration is shown in Fig.~\ref{fig:ga}, supposing there is only one step in the iterative method.

\textbf{Pre-training domain:} Because the gradients of UAP obtained by Eq.~\eqref{eq:UAP} represent the directions to the closest points on the decision boundary in the pre-training domain. Thus, limiting the feasible zone to  $\Delta\vect{\delta}_{\vect{\theta},\mathcal{D}_p}^{*}$ does little harm to the performance when evaluated in the pre-training domain. Meanwhile, according to the optimal objective, L4A finds the next best directions which are worse than those of UAPs. Thus, $\vect{\delta}_{\Delta\vect{\delta}_{\vect{\theta},\mathcal{D}_p}}^{*}$ performs better than $\vect{\delta}_{\Delta\vect{\delta}_{\vect{\theta},\mathcal{D}_p}}$ in the pre-training domain.

\textbf{Fine-tuning domain:} According to the fact that UAP bears a low gradient alignment near to 0, the subspace spanned by the tensors in  $\Delta\vect{\delta}_{\vect{\Delta}_{\vect{\theta},\mathcal{D}_p}}^{*}$ is almost orthogonal to that spanned by the tensors in $\Delta\vect{\delta}_{\vect{\Delta}_{\vect{\theta},\mathcal{D}_t}}^{*}$ which represent the best directions that send the sample to the decision boundary in the fine-tuning domain. Thus, limiting the feasible zone of Eq.~\eqref{optimization} to  $\Delta\vect{\delta}_{\vect{\theta},\mathcal{D}_p}^{*}$ suffers a great drop in ASR when evaluated on the downstream tasks in Eq.~\eqref{reformulation}.
However, as shown in Table~\ref{tab:ga}, our algorithm can achieve a gradient alignment of up to 0.6125, which means that there is considerable overlap in the next best feasible region of $\mathcal{D}_p$ in the pre-training domain obtained by L4A and that of $\mathcal{D}_f$ in the fine-tuning domain. Thus the performance of the best solution in $\Delta\vect{\delta}_{\vect{\theta},\mathcal{D}_p}$ is close to that of $\Delta\vect{\delta}_{\vect{\theta},\mathcal{D}_f}$, which represents the next best solution in the fine-tuning domain. Finally we have $\vect{\delta}_{\Delta\vect{\delta}_{\vect{\theta},\mathcal{D}_p}}$ performs better than $\vect{\delta}_{\Delta\vect{\delta}_{\vect{\theta},\mathcal{D}_p}}^{*}$ in the fine-tuning domain. 

In conclusion, \emph{the high gradient alignment guarantees high cross-finetuning transferability}.

\section{Societal impact}
A potential negative societal impact of L4A is that malicious adversaries could use it to cause security/safety issues in real-world applications. As more people focus on the pre-trained models because of their excellent performance, fine-tuning pre-trained models provided by the cloud server becomes a panacea for deep learning practitioners. In such settings, PAPs become a significant security flaw--as one can easily access the prototype pre-trained models and perform attacking algorithms on them. Our work appeals to big companies to delve further into the safety problem related to the vulnerability of pre-trained models.

\section{Conclusion}\label{sec:conclusion}

In this paper, we address the safety problem of pre-trained models. In particular, an attacker can use them to generate so-called pre-trained adversarial perturbations, achieving a high success rate on the fine-tuned models without knowing the victim model and the specific downstream tasks. Considering the inner qualities of the pre-training to fine-tuning paradigm, we propose a novel algorithm, L4A, which performs well in such problem settings. A limitation of L4A is that it performs worse than UAPs in the pre-training domain; we hope some upcoming work can fill the gap. Furthermore, L4A only utilizes the information in the pre-training domain. When the attacker obtains some information about the downstream tasks, like several unlabeled instances in the fine-tuning domain, he may be able to enhance PAPs using the knowledge and further exacerbate the situation, which we leave to future work. Thus, we hope our work can draw attention to the safety problem of pre-trained models to guarantee security.

\section*{Acknowledgement}

This work was supported by the National Key Research and Development Program of China (2020AAA0106000, 2020AAA0104304, 2020AAA0106302), NSFC Projects (Nos. 62061136001, 62076145, 62076147, U19B2034, U1811461, U19A2081, 61972224), Beijing NSF Project (No. JQ19016), BNRist (BNR2022RC01006), Tsinghua Institute for Guo Qiang, and the High Performance Computing Center, Tsinghua University. Y. Dong was also supported by the China National Postdoctoral Program for Innovative Talents and Shuimu Tsinghua Scholar Program.

\bibliographystyle{plainnat}
\bibliography{contrastive.bib}

\begin{thebibliography}{64}
\providecommand{\natexlab}[1]{#1}
\providecommand{\url}[1]{\texttt{#1}}
\expandafter\ifx\csname urlstyle\endcsname\relax
  \providecommand{\doi}[1]{doi: #1}\else
  \providecommand{\doi}{doi: \begingroup \urlstyle{rm}\Url}\fi

\bibitem[Bossard et~al.(2014)Bossard, Guillaumin, and Gool]{bossard2014food}
Lukas Bossard, Matthieu Guillaumin, and Luc~Van Gool.
\newblock Food-101--mining discriminative components with random forests.
\newblock In \emph{European Conference on Computer Vision (ECCV)}, pages
  446--461, 2014.

\bibitem[Brown et~al.(2020)Brown, Mann, Ryder, Subbiah, Kaplan, Dhariwal,
  Neelakantan, Shyam, Sastry, Askell, et~al.]{brown2020language}
Tom Brown, Benjamin Mann, Nick Ryder, Melanie Subbiah, Jared~D Kaplan, Prafulla
  Dhariwal, Arvind Neelakantan, Pranav Shyam, Girish Sastry, Amanda Askell,
  et~al.
\newblock Language models are few-shot learners.
\newblock In \emph{Advances in Neural Information Processing Systems
  (NeurIPS)}, pages 1877--1901, 2020.

\bibitem[Chen et~al.(2020{\natexlab{a}})Chen, Liu, Chang, Cheng, Amini, and
  Wang]{chen2020adversarial}
Tianlong Chen, Sijia Liu, Shiyu Chang, Yu~Cheng, Lisa Amini, and Zhangyang
  Wang.
\newblock Adversarial robustness: From self-supervised pre-training to
  fine-tuning.
\newblock In \emph{Proceedings of the IEEE/CVF Conference on Computer Vision
  and Pattern Recognition (CVPR)}, pages 699--708, 2020{\natexlab{a}}.

\bibitem[Chen et~al.(2020{\natexlab{b}})Chen, Kornblith, Norouzi, and
  Hinton]{chen2020simple}
Ting Chen, Simon Kornblith, Mohammad Norouzi, and Geoffrey Hinton.
\newblock A simple framework for contrastive learning of visual
  representations.
\newblock In \emph{International Conference on Machine Learning (ICML)}, pages
  1597--1607, 2020{\natexlab{b}}.

\bibitem[Chen et~al.(2020{\natexlab{c}})Chen, Kornblith, Swersky, Norouzi, and
  Hinton]{chen2020big}
Ting Chen, Simon Kornblith, Kevin Swersky, Mohammad Norouzi, and Geoffrey~E
  Hinton.
\newblock Big self-supervised models are strong semi-supervised learners.
\newblock In \emph{Advances in Neural Information Processing Systems
  (NeurIPS)}, pages 22243--22255, 2020{\natexlab{c}}.

\bibitem[Chen and Ma(2021)]{chen2021towards}
Tong Chen and Zhan Ma.
\newblock Towards robust neural image compression: Adversarial attack and model
  finetuning.
\newblock \emph{arXiv preprint arXiv:2112.08691}, 2021.

\bibitem[Cimpoi et~al.(2014)Cimpoi, Maji, Kokkinos, Mohamed, and
  Vedaldi]{cimpoi2014DTD}
Mircea Cimpoi, Subhransu Maji, Iasonas Kokkinos, Sammy Mohamed, and Andrea
  Vedaldi.
\newblock Describing textures in the wild.
\newblock In \emph{Proceedings of the IEEE Conference on Computer Vision and
  Pattern Recognition (CVPR)}, pages 3606--3613, 2014.

\bibitem[Coates et~al.(2011)Coates, Ng, and Lee]{coates2011stl10}
Adam Coates, Andrew Ng, and Honglak Lee.
\newblock An analysis of single-layer networks in unsupervised feature
  learning.
\newblock In \emph{Proceedings of the Fourteenth International Conference on
  Artificial Intelligence and Statistics (AISTATS)}, pages 215--223, 2011.

\bibitem[Deng and Karam(2020)]{deng2020uapepgd}
Yingpeng Deng and Lina~J Karam.
\newblock Universal adversarial attack via enhanced projected gradient descent.
\newblock In \emph{IEEE International Conference on Image Processing (ICIP)},
  pages 1241--1245, 2020.

\bibitem[Dong et~al.(2021)Dong, Luu, Lin, Yan, and Zhang]{dong2021should}
Xinshuai Dong, Anh~Tuan Luu, Min Lin, Shuicheng Yan, and Hanwang Zhang.
\newblock How should pre-trained language models be fine-tuned towards
  adversarial robustness?
\newblock In \emph{Advances in Neural Information Processing Systems
  (NeurIPS)}, pages 4356--4369, 2021.

\bibitem[Dong et~al.(2018)Dong, Liao, Pang, Su, Zhu, Hu, and
  Li]{dong2018boosting}
Yinpeng Dong, Fangzhou Liao, Tianyu Pang, Hang Su, Jun Zhu, Xiaolin Hu, and
  Jianguo Li.
\newblock Boosting adversarial attacks with momentum.
\newblock In \emph{Proceedings of the IEEE Conference on Computer Vision and
  Pattern Recognition (CVPR)}, pages 9185--9193, 2018.

\bibitem[Dong et~al.(2019)Dong, Pang, Su, and Zhu]{dong2019evading}
Yinpeng Dong, Tianyu Pang, Hang Su, and Jun Zhu.
\newblock Evading defenses to transferable adversarial examples by
  translation-invariant attacks.
\newblock In \emph{Proceedings of the IEEE/CVF Conference on Computer Vision
  and Pattern Recognition (CVPR)}, pages 4312--4321, 2019.

\bibitem[Fan et~al.(2021)Fan, Liu, Chen, Zhang, and Gan]{fan2021does}
Lijie Fan, Sijia Liu, Pin-Yu Chen, Gaoyuan Zhang, and Chuang Gan.
\newblock When does contrastive learning preserve adversarial robustness from
  pretraining to finetuning?
\newblock In \emph{Advances in Neural Information Processing Systems
  (NeurIPS)}, pages 21480--21492, 2021.

\bibitem[Gidaris et~al.(2018)Gidaris, Singh, and
  Komodakis]{gidaris2018unsupervisedrotation}
Spyros Gidaris, Praveer Singh, and Nikos Komodakis.
\newblock Unsupervised representation learning by predicting image rotations.
\newblock In \emph{International Conference on Learning Representations
  (ICLR)}, 2018.

\bibitem[Goodfellow et~al.(2015)Goodfellow, Shlens, and
  Szegedy]{goodfellow2014adver}
Ian~J Goodfellow, Jonathon Shlens, and Christian Szegedy.
\newblock Explaining and harnessing adversarial examples.
\newblock In \emph{International Conference on Learning Representations
  (ICLR)}, 2015.

\bibitem[Guo et~al.(2019)Guo, Shi, Kumar, Grauman, Rosing, and
  Feris]{guo2019spottune}
Yunhui Guo, Honghui Shi, Abhishek Kumar, Kristen Grauman, Tajana Rosing, and
  Rogerio Feris.
\newblock Spottune: transfer learning through adaptive fine-tuning.
\newblock In \emph{Proceedings of the IEEE/CVF Conference on Computer Vision
  and Pattern Recognition (CVPR)}, pages 4805--4814, 2019.

\bibitem[Han et~al.(2021)Han, Zhang, Ding, Gu, Liu, Huo, Qiu, Yao, Zhang,
  Zhang, et~al.]{han2021pre}
Xu~Han, Zhengyan Zhang, Ning Ding, Yuxian Gu, Xiao Liu, Yuqi Huo, Jiezhong Qiu,
  Yuan Yao, Ao~Zhang, Liang Zhang, et~al.
\newblock Pre-trained models: Past, present and future.
\newblock \emph{AI Open}, pages 225--250, 2021.

\bibitem[He et~al.(2016)He, Zhang, Ren, and Sun]{he2016resnet}
Kaiming He, Xiangyu Zhang, Shaoqing Ren, and Jian Sun.
\newblock Deep residual learning for image recognition.
\newblock In \emph{Proceedings of the IEEE Conference on Computer Vision and
  Pattern Recognition (CVPR)}, pages 770--778, 2016.

\bibitem[He et~al.(2020{\natexlab{a}})He, Fan, Wu, Xie, and
  Girshick]{he2020moco}
Kaiming He, Haoqi Fan, Yuxin Wu, Saining Xie, and Ross Girshick.
\newblock Momentum contrast for unsupervised visual representation learning.
\newblock In \emph{Proceedings of the IEEE/CVF Conference on Computer Vision
  and Pattern Recognition (CVPR)}, pages 9729--9738, 2020{\natexlab{a}}.

\bibitem[He et~al.(2020{\natexlab{b}})He, Fan, Wu, Xie, and
  Girshick]{he2020momentum}
Kaiming He, Haoqi Fan, Yuxin Wu, Saining Xie, and Ross Girshick.
\newblock Momentum contrast for unsupervised visual representation learning.
\newblock In \emph{Proceedings of the IEEE/CVF Conference on Computer Vision
  and Pattern Recognition (CVPR)}, pages 9729--9738, 2020{\natexlab{b}}.

\bibitem[He et~al.(2022)He, Chen, Xie, Li, Doll\'ar, and
  Girshick]{he2021masked}
Kaiming He, Xinlei Chen, Saining Xie, Yanghao Li, Piotr Doll\'ar, and Ross
  Girshick.
\newblock Masked autoencoders are scalable vision learners.
\newblock In \emph{Proceedings of the IEEE/CVF Conference on Computer Vision
  and Pattern Recognition (CVPR)}, pages 16000--16009, 2022.

\bibitem[Inoue(2018)]{inoue2018data}
Hiroshi Inoue.
\newblock Data augmentation by pairing samples for images classification.
\newblock \emph{arXiv preprint arXiv:1801.02929}, 2018.

\bibitem[Ioffe and Szegedy(2015)]{ioffe2015batch}
Sergey Ioffe and Christian Szegedy.
\newblock Batch normalization: Accelerating deep network training by reducing
  internal covariate shift.
\newblock In \emph{International Conference on Machine Learning (ICML)}, pages
  448--456, 2015.

\bibitem[Jiang et~al.(2020)Jiang, Chen, Chen, and Wang]{jiang2020robust}
Ziyu Jiang, Tianlong Chen, Ting Chen, and Zhangyang Wang.
\newblock Robust pre-training by adversarial contrastive learning.
\newblock In \emph{Advances in Neural Information Processing Systems
  (NeurIPS)}, pages 16199--16210, 2020.

\bibitem[Kenton and Toutanova(2019)]{devlin2018bert}
Jacob Devlin Ming-Wei~Chang Kenton and Lee~Kristina Toutanova.
\newblock Bert: Pre-training of deep bidirectional transformers for language
  understanding.
\newblock In \emph{Annual Conference of the North American Chapter of the
  Association for Computational Linguistics (NAACL)}, pages 4171--4186, 2019.

\bibitem[Khosla et~al.(2020)Khosla, Teterwak, Wang, Sarna, Tian, Isola,
  Maschinot, Liu, and Krishnan]{khosla2020supervised}
Prannay Khosla, Piotr Teterwak, Chen Wang, Aaron Sarna, Yonglong Tian, Phillip
  Isola, Aaron Maschinot, Ce~Liu, and Dilip Krishnan.
\newblock Supervised contrastive learning.
\newblock In \emph{Advances in Neural Information Processing Systems
  (NeurIPS)}, pages 18661--18673, 2020.

\bibitem[Khrulkov and Oseledets(2018)]{khrulkov2018asv}
Valentin Khrulkov and Ivan Oseledets.
\newblock Art of singular vectors and universal adversarial perturbations.
\newblock In \emph{Proceedings of the IEEE Conference on Computer Vision and
  Pattern Recognition (CVPR)}, pages 8562--8570, 2018.

\bibitem[Krause et~al.(2013)Krause, Deng, Stark, and Fei-Fei]{krause2013cars}
Jonathan Krause, Jia Deng, Michael Stark, and Li~Fei-Fei.
\newblock Collecting a large-scale dataset of fine-grained cars.
\newblock 2013.

\bibitem[Krizhevsky et~al.(2009)Krizhevsky, Hinton,
  et~al.]{krizhevsky2009cifar100}
Alex Krizhevsky, Geoffrey Hinton, et~al.
\newblock Learning multiple layers of features from tiny images.
\newblock 2009.

\bibitem[Kumar et~al.(2022)Kumar, Raghunathan, Jones, Ma, and
  Liang]{kumar2022fine}
Ananya Kumar, Aditi Raghunathan, Robbie Jones, Tengyu Ma, and Percy Liang.
\newblock Fine-tuning can distort pretrained features and underperform
  out-of-distribution.
\newblock \emph{The International Conference on Learning Representations
  (ICLR)}, 2022.

\bibitem[Kurakin et~al.(2018)Kurakin, Goodfellow, and
  Bengio]{kurakin2018physical}
Alexey Kurakin, Ian~J Goodfellow, and Samy Bengio.
\newblock Adversarial examples in the physical world.
\newblock In \emph{Artificial Intelligence Safety and Security}, pages 99--112.
  2018.

\bibitem[Le-Khac et~al.(2020)Le-Khac, Healy, and Smeaton]{le2020contrastive}
Phuc~H Le-Khac, Graham Healy, and Alan~F Smeaton.
\newblock Contrastive representation learning: A framework and review.
\newblock \emph{IEEE Access}, pages 193907--193934, 2020.

\bibitem[Liu et~al.(2017)Liu, Chen, Liu, and Song]{liu2016blackbox}
Yanpei Liu, Xinyun Chen, Chang Liu, and Dawn Song.
\newblock Delving into transferable adversarial examples and black-box attacks.
\newblock In \emph{International Conference on Learning Representations
  (ICLR)}, 2017.

\bibitem[Liu et~al.(2019)Liu, Ott, Goyal, Du, Joshi, Chen, Levy, Lewis,
  Zettlemoyer, and Stoyanov]{liu2019roberta}
Yinhan Liu, Myle Ott, Naman Goyal, Jingfei Du, Mandar Joshi, Danqi Chen, Omer
  Levy, Mike Lewis, Luke Zettlemoyer, and Veselin Stoyanov.
\newblock Roberta: A robustly optimized bert pretraining approach.
\newblock \emph{arXiv preprint arXiv:1907.11692}, 2019.

\bibitem[Lu et~al.(2020{\natexlab{a}})Lu, Jia, Wang, Li, Chai, Carin, and
  Velipasalar]{lu2020DR}
Yantao Lu, Yunhan Jia, Jianyu Wang, Bai Li, Weiheng Chai, Lawrence Carin, and
  Senem Velipasalar.
\newblock Enhancing cross-task black-box transferability of adversarial
  examples with dispersion reduction.
\newblock In \emph{Proceedings of the IEEE/CVF Conference on Computer Vision
  and Pattern Recognition (CVPR)}, pages 940--949, 2020{\natexlab{a}}.

\bibitem[Lu et~al.(2020{\natexlab{b}})Lu, Jia, Wang, Li, Chai, Carin, and
  Velipasalar]{lu2020std}
Yantao Lu, Yunhan Jia, Jianyu Wang, Bai Li, Weiheng Chai, Lawrence Carin, and
  Senem Velipasalar.
\newblock Enhancing cross-task black-box transferability of adversarial
  examples with dispersion reduction.
\newblock In \emph{Proceedings of the IEEE/CVF Conference on Computer Vision
  and Pattern Recognition (CVPR)}, pages 940--949, 2020{\natexlab{b}}.

\bibitem[Madry et~al.(2018)Madry, Makelov, Schmidt, Tsipras, and
  Vladu]{madry2017PGD}
Aleksander Madry, Aleksandar Makelov, Ludwig Schmidt, Dimitris Tsipras, and
  Adrian Vladu.
\newblock Towards deep learning models resistant to adversarial attacks.
\newblock In \emph{International Conference on Learning Representations
  (ICLR)}, 2018.

\bibitem[Maji et~al.(2013)Maji, Rahtu, Kannala, Blaschko, and
  Vedaldi]{maji2013fgvc}
Subhransu Maji, Esa Rahtu, Juho Kannala, Matthew Blaschko, and Andrea Vedaldi.
\newblock Fine-grained visual classification of aircraft.
\newblock \emph{arXiv preprint arXiv:1306.5151}, 2013.

\bibitem[Moosavi-Dezfooli et~al.(2016)Moosavi-Dezfooli, Fawzi, and
  Frossard]{moosavi2016deepfool}
Seyed-Mohsen Moosavi-Dezfooli, Alhussein Fawzi, and Pascal Frossard.
\newblock Deepfool: a simple and accurate method to fool deep neural networks.
\newblock In \emph{Proceedings of the IEEE conference on computer vision and
  pattern recognition (CVPR)}, pages 2574--2582, 2016.

\bibitem[Moosavi-Dezfooli et~al.(2017)Moosavi-Dezfooli, Fawzi, Fawzi, and
  Frossard]{moosavi2017universal}
Seyed-Mohsen Moosavi-Dezfooli, Alhussein Fawzi, Omar Fawzi, and Pascal
  Frossard.
\newblock Universal adversarial perturbations.
\newblock In \emph{Proceedings of the IEEE Conference on Computer Vision and
  Pattern Recognition (CVPR)}, pages 1765--1773, 2017.

\bibitem[Mopuri et~al.(2017)Mopuri, Garg, and Babu]{mopuri2017fff}
Konda~Reddy Mopuri, Utsav Garg, and R~Venkatesh Babu.
\newblock Fast feature fool: A data independent approach to universal
  adversarial perturbations.
\newblock In \emph{British Machine Vision Conference (BMVC)}, 2017.

\bibitem[Naseer et~al.(2019)Naseer, Khan, Khan, Shahbaz~Khan, and
  Porikli]{naseer2019cross}
Muhammad~Muzammal Naseer, Salman~H Khan, Muhammad~Haris Khan, Fahad
  Shahbaz~Khan, and Fatih Porikli.
\newblock Cross-domain transferability of adversarial perturbations.
\newblock In \emph{Advances in Neural Information Processing Systems
  (NeurIPS)}, pages 12905--12915, 2019.

\bibitem[Naseer et~al.(2020)Naseer, Khan, Hayat, Khan, and
  Porikli]{naseer2020ssp}
Muzammal Naseer, Salman Khan, Munawar Hayat, Fahad~Shahbaz Khan, and Fatih
  Porikli.
\newblock A self-supervised approach for adversarial robustness.
\newblock In \emph{Proceedings of the IEEE/CVF Conference on Computer Vision
  and Pattern Recognition (CVPR)}, pages 262--271, 2020.

\bibitem[Netzer et~al.(2011)Netzer, Wang, Coates, Bissacco, Wu, and
  Ng]{netzer2011svhn}
Yuval Netzer, Tao Wang, Adam Coates, Alessandro Bissacco, Bo~Wu, and Andrew~Y
  Ng.
\newblock Reading digits in natural images with unsupervised feature learning.
\newblock 2011.

\bibitem[Noroozi and Favaro(2016)]{noroozi2016jig}
Mehdi Noroozi and Paolo Favaro.
\newblock Unsupervised learning of visual representations by solving jigsaw
  puzzles.
\newblock In \emph{European Conference on Computer Vision (ECCV)}, pages
  69--84, 2016.

\bibitem[Oquab et~al.(2014)Oquab, Bottou, Laptev, and Sivic]{oquab2014transfer}
Maxime Oquab, Leon Bottou, Ivan Laptev, and Josef Sivic.
\newblock Learning and transferring mid-level image representations using
  convolutional neural networks.
\newblock In \emph{Proceedings of the IEEE Conference on Computer Vision and
  Pattern Recognition (CVPR)}, pages 1717--1724, 2014.

\bibitem[Papernot et~al.(2016)Papernot, McDaniel, Jha, Fredrikson, Celik, and
  Swami]{papernot2016lJMSA}
Nicolas Papernot, Patrick McDaniel, Somesh Jha, Matt Fredrikson, Z~Berkay
  Celik, and Ananthram Swami.
\newblock The limitations of deep learning in adversarial settings.
\newblock In \emph{2016 IEEE European Symposium on Security and Privacy
  (EuroS\&P)}, pages 372--387, 2016.

\bibitem[Park et~al.(2020)Park, Efros, Zhang, and Zhu]{park2020contrastive}
Taesung Park, Alexei~A Efros, Richard Zhang, and Jun-Yan Zhu.
\newblock Contrastive learning for unpaired image-to-image translation.
\newblock In \emph{European Conference on Computer Vision (ECCV)}, pages
  319--345, 2020.

\bibitem[Parkhi et~al.(2012)Parkhi, Vedaldi, Zisserman, and
  Jawahar]{parkhi2012pets}
Omkar~M Parkhi, Andrea Vedaldi, Andrew Zisserman, and CV~Jawahar.
\newblock Cats and dogs.
\newblock In \emph{Proceedings of the IEEE Conference on Computer Vision and
  Pattern Recognition (CVPR)}, pages 3498--3505, 2012.

\bibitem[Qiu et~al.(2020)Qiu, Sun, Xu, Shao, Dai, and Huang]{qiu2020pre}
Xipeng Qiu, Tianxiang Sun, Yige Xu, Yunfan Shao, Ning Dai, and Xuanjing Huang.
\newblock Pre-trained models for natural language processing: A survey.
\newblock \emph{Science China Technological Sciences}, pages 1872--1897, 2020.

\bibitem[Radford et~al.(2021)Radford, Kim, Hallacy, Ramesh, Goh, Agarwal,
  Sastry, Askell, Mishkin, Clark, et~al.]{radford2021clip}
Alec Radford, Jong~Wook Kim, Chris Hallacy, Aditya Ramesh, Gabriel Goh,
  Sandhini Agarwal, Girish Sastry, Amanda Askell, Pamela Mishkin, Jack Clark,
  et~al.
\newblock Learning transferable visual models from natural language
  supervision.
\newblock In \emph{International Conference on Machine Learning (ICML)}, pages
  8748--8763, 2021.

\bibitem[Russakovsky et~al.(2015)Russakovsky, Deng, Su, Krause, Satheesh, Ma,
  Huang, Karpathy, Khosla, Bernstein, et~al.]{russakovsky2015imagenet}
Olga Russakovsky, Jia Deng, Hao Su, Jonathan Krause, Sanjeev Satheesh, Sean Ma,
  Zhiheng Huang, Andrej Karpathy, Aditya Khosla, Michael Bernstein, et~al.
\newblock Imagenet large scale visual recognition challenge.
\newblock \emph{International journal of computer vision (IJCV)}, pages
  211--252, 2015.

\bibitem[Sai et~al.(2020)Sai, Mohankumar, Arora, and Khapra]{sai2020improving}
Ananya~B Sai, Akash~Kumar Mohankumar, Siddhartha Arora, and Mitesh~M Khapra.
\newblock Improving dialog evaluation with a multi-reference adversarial
  dataset and large scale pretraining.
\newblock \emph{Transactions of the Association for Computational Linguistics
  (TACL)}, pages 810--827, 2020.

\bibitem[Szegedy et~al.(2014)Szegedy, Zaremba, Sutskever, Bruna, Erhan,
  Goodfellow, and Fergus]{szegedy2013FGSM}
Christian Szegedy, Wojciech Zaremba, Ilya Sutskever, Joan Bruna, Dumitru Erhan,
  Ian Goodfellow, and Rob Fergus.
\newblock Intriguing properties of neural networks.
\newblock In \emph{International Conference on Learning Representations
  (ICLR)}, 2014.

\bibitem[Trinh et~al.(2019)Trinh, Luong, and Le]{trinh2019selfie}
Trieu~H Trinh, Minh-Thang Luong, and Quoc~V Le.
\newblock Selfie: Self-supervised pretraining for image embedding.
\newblock \emph{arXiv preprint arXiv:1906.02940}, 2019.

\bibitem[Vaswani et~al.(2017)Vaswani, Shazeer, Parmar, Uszkoreit, Jones, Gomez,
  Kaiser, and Polosukhin]{vaswani2017transformer}
Ashish Vaswani, Noam Shazeer, Niki Parmar, Jakob Uszkoreit, Llion Jones,
  Aidan~N Gomez, {\L}ukasz Kaiser, and Illia Polosukhin.
\newblock Attention is all you need.
\newblock In \emph{Advances in neural information processing systems
  (NeurIPS)}, pages 6000--6010, 2017.

\bibitem[Wah et~al.(2011)Wah, Branson, Welinder, Perona, and
  Belongie]{wah2011cub}
Catherine Wah, Steve Branson, Peter Welinder, Pietro Perona, and Serge
  Belongie.
\newblock The caltech-ucsd birds-200-2011 dataset.
\newblock 2011.

\bibitem[Xie et~al.(2019)Xie, Zhang, Zhou, Bai, Wang, Ren, and
  Yuille]{xie2019improving}
Cihang Xie, Zhishuai Zhang, Yuyin Zhou, Song Bai, Jianyu Wang, Zhou Ren, and
  Alan~L Yuille.
\newblock Improving transferability of adversarial examples with input
  diversity.
\newblock In \emph{Proceedings of the IEEE/CVF Conference on Computer Vision
  and Pattern Recognition (CVPR)}, pages 2730--2739, 2019.

\bibitem[Yan et~al.(2020)Yan, Misra, Gupta, Ghadiyaram, and
  Mahajan]{yan2020clusterfit}
Xueting Yan, Ishan Misra, Abhinav Gupta, Deepti Ghadiyaram, and Dhruv Mahajan.
\newblock Clusterfit: Improving generalization of visual representations.
\newblock In \emph{Proceedings of the IEEE/CVF Conference on Computer Vision
  and Pattern Recognition (CVPR)}, pages 6509--6518, 2020.

\bibitem[Yang and Liu(2022)]{yang2022robust}
Zonghan Yang and Yang Liu.
\newblock On robust prefix-tuning for text classification.
\newblock In \emph{International Conference on Learning Representations
  (ICLR)}, 2022.

\bibitem[Yosinski et~al.(2014)Yosinski, Clune, Bengio, and
  Lipson]{yosinski2014transferable}
Jason Yosinski, Jeff Clune, Yoshua Bengio, and Hod Lipson.
\newblock How transferable are features in deep neural networks?
\newblock In \emph{Advances in Neural Information Processing Systems
  (NeurIPS)}, pages 3320--3328, 2014.

\bibitem[Yun et~al.(2019)Yun, Han, Oh, Chun, Choe, and Yoo]{yun2019cutmix}
Sangdoo Yun, Dongyoon Han, Seong~Joon Oh, Sanghyuk Chun, Junsuk Choe, and
  Youngjoon Yoo.
\newblock Cutmix: Regularization strategy to train strong classifiers with
  localizable features.
\newblock In \emph{Proceedings of the IEEE/CVF International Conference on
  Computer Vision (ICCV)}, pages 6023--6032, 2019.

\bibitem[Zhang et~al.(2021)Zhang, Li, Chen, Song, Gao, He, and
  Xue]{zhang2022beyond}
Qilong Zhang, Xiaodan Li, Yuefeng Chen, Jingkuan Song, Lianli Gao, Yuan He, and
  Hui Xue.
\newblock Beyond imagenet attack: Towards crafting adversarial examples for
  black-box domains.
\newblock In \emph{International Conference on Learning Representations
  (ICLR)}, 2021.

\bibitem[Zhang et~al.(2016)Zhang, Isola, and Efros]{2016Colorful}
Richard Zhang, Phillip Isola, and Alexei~A Efros.
\newblock Colorful image colorization.
\newblock In \emph{European Conference on Computer Vision (ECCV)}, pages
  649--666, 2016.

\end{thebibliography}

\clearpage
\section*{Checklist}
\begin{enumerate}

\item For all authors...
\begin{enumerate}
  \item Do the main claims made in the abstract and introduction accurately reflect the paper's contributions and scope?
    \answerYes{}
  \item Did you describe the limitations of your work?
    \answerYes{See Section~\ref{sec:conclusion}.}
  \item Did you discuss any potential negative societal impacts of your work?
    \answerYes{See Section~\ref{sec:conclusion}.}
  \item Have you read the ethics review guidelines and ensured that your paper conforms to them?
    \answerYes{}
\end{enumerate}

\item If you are including theoretical results...
\begin{enumerate}
  \item Did you state the full set of assumptions of all theoretical results?
    \answerNA{}
        \item Did you include complete proofs of all theoretical results?
    \answerNA{}
\end{enumerate}

\item If you ran experiments...
\begin{enumerate}
  \item Did you include the code, data, and instructions needed to reproduce the main experimental results (either in the supplemental material or as a URL)?
    \answerYes{See Appendix.}
  \item Did you specify all the training details (e.g., data splits, hyperparameters, how they were chosen)?
    \answerYes{See Section~\ref{sec:experiment} and Appendix {\color{red} F}.}
        \item Did you report error bars (e.g., with respect to the random seed after running experiments multiple times)?
    \answerNo{Since running L4A on all the ten dataset and the three models is time-consuming, we only ran one time.}
        \item Did you include the total amount of compute and the type of resources used (e.g., type of GPUs, internal cluster, or cloud provider)?
    \answerYes{See Appendix {\color{red} F}.}
\end{enumerate}

\item If you are using existing assets (e.g., code, data, models) or curating/releasing new assets...
\begin{enumerate}
  \item If your work uses existing assets, did you cite the creators?
    \answerYes{}
  \item Did you mention the license of the assets?
    \answerYes{See Appendix {\color{red} D}.}
  \item Did you include any new assets either in the supplemental material or as a URL?
    \answerYes{See supplementary material.}
  \item Did you discuss whether and how consent was obtained from people whose data you're using/curating?
    \answerYes{See Appendix {\color{red} D}.}
  \item Did you discuss whether the data you are using/curating contains personally identifiable information or offensive content?
    \answerYes{See Appendix {\color{red} D}.}
\end{enumerate}

\item If you used crowdsourcing or conducted research with human subjects...
\begin{enumerate}
  \item Did you include the full text of instructions given to participants and screenshots, if applicable?
    \answerNA{}
  \item Did you describe any potential participant risks, with links to Institutional Review Board (IRB) approvals, if applicable?
    \answerNA{}
  \item Did you include the estimated hourly wage paid to participants and the total amount spent on participant compensation?
    \answerNA{}
\end{enumerate}

\end{enumerate}

\clearpage
\appendix

\section{Additional experiments}
\subsection{Additional experiments on other pre-trained models}\label{appaddpre}
In this section, we report the results on CLIP and MOCO in Table~\ref{tab:moco} and Table~\ref{tab:clip}, respectively. Note that the first seven columns of validation datasets are fine-grained, while the next three are coarse-grained ones. We mark the best results for each dataset in \textbf{bold}, and the best baseline in \textcolor{blue}{\textbf{blue}}. We highlight the results of L4A\textsubscript{ugs} in \textcolor{red}{red} to emphasize that the Low-Lever Layer Lifting attack equipped with Uniform Gaussian Sampling shows great \emph{cross-finetuning transferability} and performs best.\\
These tables show that our proposed methods outperform all the baselines by a large margin. For example, as can be seen from Table~\ref{tab:moco}, if the target model is Resnet50 pre-trained by MOCO, the best competitor UAP achieves an average attack success rate of 54.34\%, while the L4A\textsubscript{fuse} can lift it up to \textbf{54.72\%} and the UGS technique further boosts the performance up to \textbf{59.72\%}. 

\begin{table}[H]\footnotesize
\setlength{\tabcolsep}{5pt}
\caption{The attack success rate(\%) of various methods we study against \textbf{Resnet50} pretrained by \textbf{MOCO}. Note that C10 stands for CIFAR10, and C100 stands for CIFAR100.}

\begin{tabular}{l||ccccccc|ccc|c}
\hline
ASR
& Cars                                    & Pets                                    & Food                           & DTD                                     & FGVC                                    & CUB                                     & SVHN                                    & C10                                 & C100                                & STL10                                   & AVG                                     \\ \hline\hline
FFF\textsubscript{no}                         & 30.72                                 & 25.59                                 & 60.03                                 & 48.51                                 & 84.97                                 & 62.82                        & 4.92                        & 17.22                        & 55.37                                 & 12.84                                 & 40.30                                 \\
FFF\textsubscript{mean}                       & 39.04                                 & 32.49                                 & 68.94                                 & 52.93                                 & 85.57                                 & \textbf{68.93}               & 8.02                        & 22.23                        & 67.53                                 & 14.26                                 & 45.99                                 \\
FFF\textsubscript{one}                        & 31.82                                 & 27.26                                 & 53.06                                 & 51.38                                 & 77.92                                 & 54.42                        & 5.09                        & 17.43                        & 57.26                                 & 10.23                                 & 38.59                                 \\
DR                            & 44.30                                 & 39.63                                 & 51.11                                 & 53.51                                 & 81.43                                 & 55.85                        & 5.01                        & 30.24                        & 74.25                                 & 13.91                                 & 44.92                                 \\
SSP                           & 33.75                                 & 36.44                                 & 75.32                                 & 60.00                                 & 80.32                                 & 68.73                        & 5.90                        & 25.42                        & 69.27                                 & 29.03                                 & 48.42                                 \\
ASV                           & 40.01                                 & 32.27                                 & 60.63                                 & 47.50                                 & 82.87                                 & 62.77                        & \textbf{41.71}              & 11.15                        & 50.30                                 & 9.05                                  & 43.83                                 \\
{\color[HTML]{3531FF} UAP}    & {\color[HTML]{3531FF} 61.00}          & {\color[HTML]{3531FF} 52.44}          & {\color[HTML]{3531FF} 77.00}          & {\color[HTML]{3531FF} 60.75}          & {\color[HTML]{3531FF} 83.79}          & {\color[HTML]{3531FF} 68.54} & {\color[HTML]{3531FF} 5.75} & {\color[HTML]{3531FF} 28.02} & {\color[HTML]{3531FF} 68.22}          & {\color[HTML]{3531FF} 37.94}          & {\color[HTML]{3531FF} 54.34}          \\
UAPEPGD                       & 45.55                                 & 38.05                                 & 70.12                                 & 54.79                                 & 69.04                                 & 57.25                        & 3.82                        & 10.47                        & 48.38                                 & 21.44                                 & 41.89                                 \\ \hline
L4A\textsubscript{base}                       & 44.10                                 & 51.86                                 & 77.44                                 & 62.61                                 & 81.49                                 & 61.30                        & 5.65                        & 45.70                        & 81.88                                 & 27.33                                 & 53.94                                 \\
L4A\textsubscript{fuse}                       & 44.25                                 & 54.02                                 & 78.09                                 & 63.19                                 & 82.90                                 & 63.26                        & 5.13                        & \textbf{46.71}               & 81.66                                 & 27.95                                 & 54.72                                 \\
{\color[HTML]{FE0000} L4A\textsubscript{ugs}} & {\color[HTML]{FE0000} \textbf{61.22}} & {\color[HTML]{FE0000} \textbf{58.11}} & {\color[HTML]{FE0000} \textbf{86.52}} & {\color[HTML]{FE0000} \textbf{67.71}} & {\color[HTML]{FE0000} \textbf{88.96}} & {\color[HTML]{FE0000} 65.57} & {\color[HTML]{FE0000} 5.08} & {\color[HTML]{FE0000} 39.23} & {\color[HTML]{FE0000} \textbf{83.12}} & {\color[HTML]{FE0000} \textbf{41.93}} & {\color[HTML]{FE0000} \textbf{59.74}} \\ \hline
\end{tabular}\label{tab:moco}
\centering
\end{table}

\begin{table}[H]\footnotesize
\setlength{\tabcolsep}{5pt}
\caption{The attack success rate(\%) of various methods we study against \textbf{Resnet50} pretrained by \textbf{CLIP}. Note that C10 stands for CIFAR10, and C100 stands for CIFAR100.}
\begin{tabular}{l||ccccccc|ccc|c}
\hline
ASR
& Cars                                    & Pets                                    & Food                           & DTD                                     & FGVC                                    & CUB                                     & SVHN                                  & C10                                 & C100                                & STL10                                   & AVG                                     \\ \hline\hline
FFF\textsubscript{no}                         & 89.96                                 & 90.41                                 & 94.57                                 & 82.45                                 & 87.81                                 & \textbf{99.08}               & 98.50                                 & 78.20                                 & 80.41                        & 89.19                        & 89.06                                 \\
FFF\textsubscript{mean}                       & 92.86                                 & 89.78                                 & 93.28                                 & 81.49                                 & 89.30                                 & 99.00                        & 99.10                                 & 75.35                                 & 80.41                        & 89.71                        & 89.03                                 \\
FFF\textsubscript{one}                        & 89.21                                 & 86.73                                 & 93.47                                 & 80.32                                 & 85.63                                 & 99.03                        & 98.38                                 & 71.78                                 & 80.41                        & 84.89                        & 86.98                                 \\
DR                            & 71.84                                 & 52.74                                 & 77.07                                 & 70.32                                 & 86.69                                 & 99.01                        & 94.69                                 & 61.74                                 & 80.41                        & 84.06                        & 77.86                                 \\
SSP                           & 85.61                                 & 84.98                                 & 83.23                                 & 78.19                                 & 90.00                                 & 98.98                        & 98.50                                 & 75.80                                 & 80.41                        & 87.90                        & 86.36                                 \\
{\color[HTML]{3531FF} ASV}    & {\color[HTML]{3531FF} 91.04}          & {\color[HTML]{3531FF} 90.80}          & {\color[HTML]{3531FF} 93.98}          & {\color[HTML]{3531FF} 81.82}          & {\color[HTML]{3531FF} 88.96}          & {\color[HTML]{3531FF} 98.02} & {\color[HTML]{3531FF} 97.68}          & {\color[HTML]{3531FF} 78.75}          & {\color[HTML]{3531FF} 79.41} & {\color[HTML]{3531FF} 87.34} & {\color[HTML]{3531FF} 88.78}          \\
UAP                           & 75.84                                 & 65.84                                 & 87.12                                 & 73.40                                 & 88.08                                 & 99.00                        & 96.52                                 & 57.18                                 & 80.41                        & 87.03                        & 81.04                                 \\
UAPEPGD                       & 64.15                                 & 52.14                                 & 60.35                                 & 67.23                                 & 89.99                                 & 98.59                        & 95.14                                 & 47.05                                 & 80.41                        & 80.86                        & 73.59                                 \\ \hline
L4A\textsubscript{base}                       & 79.83                                 & 66.61                                 & 79.74                                 & 70.53                                 & 85.96                                 & 99.03                        & 98.04                                 & 64.00                                 & 84.01                        & 82.26                        & 81.00                                 \\
L4A\textsubscript{fuse}                       & 96.57                                 & 96.27                                 & 98.39                                 & 86.01                                 & 88.11                                 & 99.00                        & 98.41                                 & 80.41                                 & \textbf{80.95}               & \textbf{90.01}               & 91.41                                 \\
{\color[HTML]{FE0000} L4A\textsubscript{ugs}} & {\color[HTML]{FE0000} \textbf{97.12}} & {\color[HTML]{FE0000} \textbf{97.19}} & {\color[HTML]{FE0000} \textbf{98.67}} & {\color[HTML]{FE0000} \textbf{91.12}} & {\color[HTML]{FE0000} \textbf{91.56}} & {\color[HTML]{FE0000} 99.02} & {\color[HTML]{FE0000} \textbf{99.10}} & {\color[HTML]{FE0000} \textbf{96.55}} & {\color[HTML]{FE0000} 80.41} & {\color[HTML]{FE0000} 85.36} & {\color[HTML]{FE0000} \textbf{93.61}} \\ \hline

\end{tabular}\label{tab:clip}
\end{table}

\subsection{PAPs against adversarial fine-tuned models}
In our paper, we only conduct experiments on standard fine-tuning. We followed the training process introduced by \cite{dong2021should}, which adopts a distillation term to preserve high-quality features of the pretrained model to boost model performance from a view of information theory. We do some additional experiments on \textbf{Resnet50} pretrained by \textbf{SimCLRv2}. And the results are as follows.
\begin{table}[H]\footnotesize
\setlength{\tabcolsep}{5pt}
\caption{The attack success rate(\%) of different methods against adversarial fine-tuned models. Note that C10 stands for CIFAR10 and C100 stands for CIFAR100.}
\begin{tabular}{l||ccccccc|ccc|c}
\hline
ASR
& Cars                                    & Pets                                    & Food                           & DTD                                     & FGVC                                    & CUB                                     & SVHN                                    & C10                                 & C100                                & STL10                                   & AVG                                     \\ \hline\hline
FFF\textsubscript{no}                          & 14.69                        & 26.08                                 & 50.86                        & 48.40                                 & 27.68                        & 31.74                                 & 34.35                                 & 36.78                                 & 10.30                                 & 16.56                                 & 29.74                                 \\
FFF\textsubscript{mean}                        & \textbf{14.72}               & 26.30                                 & 50.36                        & 48.14                                 & 27.70                        & 31.90                                 & 33.93                                 & 36.74                                 & 9.97                                  & 16.61                                 & 29.64                                 \\
{\color[HTML]{3531FF} FFF\textsubscript{one}}  & {\color[HTML]{3531FF} 14.65} & {\color[HTML]{3531FF} 26.53}          & {\color[HTML]{3531FF} 53.08} & {\color[HTML]{3531FF} 48.52}          & {\color[HTML]{3531FF} 27.06} & {\color[HTML]{3531FF} 31.49}          & {\color[HTML]{3531FF} \textbf{34.02}} & {\color[HTML]{3531FF} 37.35}          & {\color[HTML]{3531FF} 10.99}          & {\color[HTML]{3531FF} 16.73}          & {\color[HTML]{3531FF} 30.04}          \\
DR                             & 14.33                        & 26.96                                 & 50.84                        & 47.13                                 & 27.39                        & 32.41                                 & 33.81                                 & 37.39                                 & 10.85                                 & 16.30                                 & 29.74                                 \\
SSP                            & 14.31                        & 26.08                                 & 50.01                        & 48.08                                 & \textbf{27.51}               & 32.10                                 & 33.14                                 & 37.14                                 & 11.20                                 & 16.34                                 & 29.59                                 \\
ASV                            & 11.95                        & 16.35                                 & 25.37                        & 37.23                                 & 24.58                        & 26.93                                 & 32.22                                 & 31.39                                 & 6.32                                  & 7.15                                  & 21.95                                 \\
UAP                            & 14.54                        & 25.67                                 & 47.92                        & 48.19                                 & 27.34                        & 32.60                                 & 33.60                                 & 36.71                                 & 10.94                                 & 16.26                                 & 29.38                                 \\
UAPEPGD                        & 14.40                        & 24.20                                 & 46.82                        & 47.18                                 & 27.32                        & 32.53                                 & 34.02                                 & 35.29                                 & 10.38                                 & 15.51                                 & 28.77                                 \\ \hline
L4A\textsubscript{base}                        & 14.48                        & 26.76                                 & \textbf{55.66}               & 50.69                                 & 26.22                        & 32.14                                 & 33.45                                 & 37.33                                 & 11.10                                 & 16.75                                 & 30.46                                 \\
{\color[HTML]{FE0000} L4Afuse} & {\color[HTML]{FE0000} 14.64} & {\color[HTML]{FE0000} \textbf{28.07}} & {\color[HTML]{FE0000} 55.59} & {\color[HTML]{FE0000} \textbf{50.74}} & {\color[HTML]{FE0000} 26.96} & {\color[HTML]{FE0000} \textbf{32.49}} & {\color[HTML]{FE0000} 33.57}          & {\color[HTML]{FE0000} \textbf{37.66}} & {\color[HTML]{FE0000} \textbf{11.31}} & {\color[HTML]{FE0000} \textbf{16.94}} & {\color[HTML]{FE0000} \textbf{30.80}} \\
L4A\textsubscript{ugs}                         & 14.35                        & 25.84                                 & 53.79                        & 50.16                                 & 26.80                        & 32.02                                 & 33.54                                 & 36.52                                 & 10.81                                 & 16.56                                 & 30.04                                 \\ \hline
\end{tabular}
\end{table}
As seen from the table, all the methods suffer degenerated performance against adversarial fine-tuning. However, L4A still performs best among these competitors. For example, considering the DTD dataset, the best baseline FFF\textsubscript{one} achieves an attack success rate of 48.52\%, while the ASR of the villain L4A\textsubscript{base} is up to 50.69\%, and the fusing loss further boosts the performance to 50.74\%. \label{adversarialfine}

\subsection{PAPs against adversarial pretrained models}
To test PAPs against adversarial pretrained models, we follow the method proposed by \cite{jiang2020robust}, which uses adversarial views to boost robustness. We first train a robust Resnet50 using enhanced SimCLRv2. Then, we generate PAPs using that model and test them on both the adversarial fine-tuned and standard fine-tuned models on downstream tasks.
In Table~\ref{tab:advpre}, we report the results on standard finetuned models.

\begin{table}[H]\footnotesize
\setlength{\tabcolsep}{5pt}
\caption{The attack success rate(\%) of different methods against adversarial-pretrained-standard-finetuned models. Note that C10 stands for CIFAR10, and C100 stands for CIFAR100.}
\begin{tabular}{l||ccccccc|ccc|c}
\hline
ASR
& Cars                                    & Pets                                    & Food                           & DTD                                     & FGVC                                    & CUB                                     & SVHN                                    & C10                                 & C100                                & STL10                                   & AVG                                     \\ \hline\hline
FFF\textsubscript{no}                          & 45.34                                 & 23.33                        & 53.16                                 & 44.95                        & 67.63                                 & 45.89                                 & 81.12                                 & 60.78                        & 95.38                        & 18.48                        & 53.61                                 \\
{\color[HTML]{3531FF} FFF\textsubscript{mean}} & {\color[HTML]{3531FF} 50.81}          & {\color[HTML]{3531FF} 27.86} & {\color[HTML]{3531FF} 60.99}          & {\color[HTML]{3531FF} 46.97} & {\color[HTML]{3531FF} 74.56}          & {\color[HTML]{3531FF} 52.69}          & {\color[HTML]{3531FF} 81.27}          & {\color[HTML]{3531FF} 71.33} & {\color[HTML]{3531FF} 99.00} & {\color[HTML]{3531FF} 28.97} & {\color[HTML]{3531FF} 59.45}          \\
FFF\textsubscript{one}                         & 44.37                                 & 25.21                        & 47.12                                 & 46.65                        & 68.35                                 & 43.89                                 & 81.50                                 & 63.36                        & 97.78                        & 15.10                        & 53.33                                 \\
DR                             & 48.81                                 & 34.15                        & 61.32                                 & 48.30                        & 79.98                                 & 51.36                                 & 76.29                                 & 63.17                        & 80.29                        & 13.22                        & 55.69                                 \\
SSP                            & 44.67                                 & \textbf{34.23}               & 44.84                                 & 48.72                        & 69.79                                 & 45.32                                 & 83.32                                 & 81.44                        & 97.95                        & 20.95                        & 54.21                                 \\
ASV                            & 47.94                                 & 25.32                        & 56.66                                 & 45.80                        & 65.94                                 & 41.14                                 & 78.52                                 & 61.25                        & 96.17                        & 12.16                        & 53.09                                 \\
UAP                            & 35.03                                 & 32.65                        & 38.39                                 & 46.44                        & 67.12                                 & 42.22                                 & 78.72                                 & 67.66                        & 93.81                        & \textbf{36.41}               & 53.84                                 \\
UAPEPGD                        & 23.20                                 & 17.85                        & 25.20                                 & 39.10                        & 53.47                                 & 31.19                                 & 66.41                                 & 13.63                        & 66.98                        & 6.33                         & 34.33                                 \\ \hline
L4A\textsubscript{base}                        & 55.93                                 & 26.41                        & 67.82                                 & 52.55                        & 79.60                                 & 56.64                                 & 84.07                                 & \textbf{84.52}               & 99.00                        & 28.16                        & 63.47                                 \\
L4A\textsubscript{fuse}                       & 67.08                                 & 25.94                        & 67.95                                 & \textbf{54.04}               & 82.20                                 & 61.41                                 & 84.05                                 & 83.09                        & \textbf{99.00}               & 33.71                        & 65.85                                 \\
{\color[HTML]{FE0000} L4A\textsubscript{ugs} }  & {\color[HTML]{FE0000} \textbf{82.46}} & {\color[HTML]{FE0000} 25.91} & {\color[HTML]{FE0000} \textbf{74.37}} & {\color[HTML]{FE0000} 51.65} & {\color[HTML]{FE0000} \textbf{88.93}} & {\color[HTML]{FE0000} \textbf{72.14}} & {\color[HTML]{FE0000} \textbf{84.08}} & {\color[HTML]{FE0000} 78.66} & {\color[HTML]{FE0000} 98.98} & {\color[HTML]{FE0000} 31.97} & {\color[HTML]{FE0000} \textbf{68.91}} \\ \hline
\end{tabular}\label{tab:advpre}
\end{table}

The above table shows that adversarial-pretrained models show little robustness after standard fine-tuning, which is also reported in \cite{chen2020adversarial,kumar2022fine}. In such settings, L4A still performs best among these competitors: the best baseline FFF\textsubscript{mean} achieves an average attack success rate of 59.45\%, while the ASR of the villain L4A\textsubscript{base} is up to 63.47\%, and the Uniform Gaussian sampling further boosts the performance to 68.91\%. Another interesting finding is that low-level-based methods, such as FFF, DR, SSP and L4A, perform better than high-level-based ones like UAPEPGD, which uses classification scores. This further supports our findings in Fig 2 and our motivation to use low-level layers.

In Table~\ref{tab:advadv}, we report the results on adversarial-finetuned models. Note that we adopt the adversarial-finetuning method in \cite{dong2021should}.

\begin{table}[H]\footnotesize
\setlength{\tabcolsep}{5pt}
\caption{The attack success rate(\%) of different methods against adversarial-pretrained-adversarial-fine-tuned models. Note that C10 stands for CIFAR10, and C100 stands for CIFAR100.}
\begin{tabular}{l||ccccccc|ccc|c}
\hline
ASR
& Cars                                    & Pets                                    & Food                           & DTD                                     & FGVC                                    & CUB                                     & \multicolumn{1}{l|}{SVHN}                                    & C10                                 & C100                                & \multicolumn{1}{l|}{STL10}                                   & AVG                                     \\ \hline\hline

FFF\textsubscript{no}                         & 14.82                        & 27.13                        & 47.84                        & 49.19                                 & 37.61                        & 38.82                        & 9.24                                  & 21.27                                 & 39.32                                 & 16.60                                 & 30.18                                 \\
FFF\textsubscript{mean}                       & 14.88                        & 27.09                        & 48.82                        & 49.40                                 & 38.01                        & 38.80                        & 9.09                                  & 20.45                                 & 39.32                                 & 16.47                                 & 30.23                                 \\
FFF\textsubscript{one}                        & 14.95                        & 27.25                        & 47.08                        & 49.40                                 & 37.80                        & 39.01                        & 9.02                                  & 21.09                                 & 39.31                                 & 16.31                                 & 30.12                                 \\
{\color[HTML]{3531FF} DR}     & {\color[HTML]{3531FF} 15.18} & {\color[HTML]{3531FF} 28.59} & {\color[HTML]{3531FF} 48.51} & {\color[HTML]{3531FF} 48.30}          & {\color[HTML]{3531FF} 38.46} & {\color[HTML]{3531FF} 38.73} & {\color[HTML]{3531FF} 8.64}           & {\color[HTML]{3531FF} 20.73}          & {\color[HTML]{3531FF} \textbf{39.61}} & {\color[HTML]{3531FF} 16.86}          & {\color[HTML]{3531FF} 30.36}          \\
SSP                           & 15.20                        & 28.67                        & 39.93                        & 46.38                                 & \textbf{38.88}               & 37.42                        & 7.35                                  & 21.89                                 & 39.49                                 & 15.11                                 & 29.03                                 \\
ASV                           & 15.56                        & 27.77                        & 49.17                        & 49.15                                 & 38.58                        & 37.83                        & 10.16                                 & 18.82                                 & 34.71                                 & 13.44                                 & 29.52                                 \\
UAP                           & 15.53                        & 27.77                        & 48.23                        & 46.96                                 & 38.55                        & 37.59                        & 7.49                                  & 15.64                                 & 35.30                                 & 14.43                                 & 28.75                                 \\
UAPEPGD                       & 15.00                        & 28.24                        & 41.39                        & 46.49                                 & 37.98                        & 37.04                        & 6.69                                  & 11.75                                 & 35.02                                 & 12.44                                 & 27.20                                 \\ \hline
L4A\textsubscript{base}                       & 15.64                        & 29.59                        & 49.18                        & 51.65                                 & 38.01                        & 39.83                        & 11.24                                 & 24.43                                 & 38.39                                 & 16.96                                 & 31.49                                 \\
L4A\textsubscript{fuse}                      & \textbf{15.90}               & \textbf{30.88}               & \textbf{49.83}               & 49.89                                 & 38.01                        & \textbf{40.05}               & 10.69                                 & 24.25                                 & 37.20                                 & 17.10                                 & 31.38                                 \\
{\color[HTML]{FE0000} L4A\textsubscript{ugs}} & {\color[HTML]{FE0000} 15.69} & {\color[HTML]{FE0000} 30.25} & {\color[HTML]{FE0000} 49.59} & {\color[HTML]{FE0000} \textbf{52.02}} & {\color[HTML]{FE0000} 37.68} & {\color[HTML]{FE0000} 39.87} & {\color[HTML]{FE0000} \textbf{11.25}} & {\color[HTML]{FE0000} \textbf{24.52}} & {\color[HTML]{FE0000} 38.06}          & {\color[HTML]{FE0000} \textbf{17.11}} & {\color[HTML]{FE0000} \textbf{31.60}} \\ \hline
\end{tabular}\label{tab:advadv}
\end{table}

From Table~\ref{tab:advadv}, we can see that adversarial-pretrained-adversarial-finetuned models show much robustness after fine-tuning, which is consistent with Appendix~\ref{adversarialfine} and the finding in \cite{chen2020adversarial,kumar2022fine} that adversarial fine-tuning contributes to the final robustness more than adversarial pre-training. Although all the methods degenerate a lot, L4A is still among the best ones.

\subsection{PAPs in other vision tasks}
We conduct experiments on semantic segmentation and object detection tasks in this subsection to evaluate our methods. 
For object detection, we adopt the off-the-shelf Resnet50 model provided by MMDetection repo, which is pre-trained by the method of MOCOv2 on ImageNet and then fine-tuned on the COCO object detection task. The results of different methods are in Table~\ref{tab:det}

\begin{table}[H]\footnotesize
\setlength{\tabcolsep}{4pt}
\caption{Objection detection finetuned on COCO. Evaluation is on COCO val2017, and results are reported in the metrics of mAP, mAP\textsubscript{50}, and mAP\textsubscript{75}. We mark the best ones for each metric in \textbf{bold}. Note that EPGD stands for the UAPEPGD method.}
\begin{tabular}{l||cccccccc|ccc}
\hline
Methods       & FFF\textsubscript{no} & FFF\textsubscript{mean} & FFF\textsubscript{one} & STD   & SSP   & {\color[HTML]{3531FF}ASV}   & UAP   & EPGD & L4A\textsubscript{base}  & L4A\textsubscript{fuse}  & {\color[HTML]{FE0000}L4A\textsubscript{ugs}} \\ \hline\hline
mAP     & 30.7    & 30.0     & 30.8       & 31.6 & 31.0 & {\color[HTML]{3531FF}29.8} & 30.2 & 34.2   & 29.8 & 29.3 & {\color[HTML]{FE0000}\textbf{26.5}}   \\
mAP\textsubscript{50} & 48.5    & 47.6     & 48.6       & 49.6 & 48.6 & {\color[HTML]{3531FF}46.9} & 47.8 & 53.1   & 46.9 & 46.2 & {\color[HTML]{FE0000}\textbf{42.5}}   \\
mAP\textsubscript{75} & 32.9    & 32.1     & 33.0       & 34.2 & 33.4 & {\color[HTML]{3531FF}31.9} & 32.5 & 37.4   & 32.0 & 31.6 & {\color[HTML]{FE0000}\textbf{28.2}}   \\ \hline
\end{tabular}\label{tab:det}
\end{table}

The table shows that our proposed methods outperform all the baselines by a large margin. For example, the best competitor ASV achieves a mAP\textsubscript{50} of 46.9\%, while the UGS technique can degenerate it to \textbf{42.5\%}, showing its effectiveness.\\

As for segmentation, we use the ViT-base model provided by MMSegmentatation, which is pre-trained by MAE on ImageNet and then finetuned on the ADE20k dataset. The results are as follows:

\begin{table}[H]\footnotesize
\setlength{\tabcolsep}{4pt}
\caption{Segmentation finetuned on ADE20k. Results are reported in the metric of mIoU. Note that EPGD stands for UAPEPGD method.}
\begin{tabular}{l||cccccccc|ccc}
\hline
methods       & {\color[HTML]{3531FF}FFF\textsubscript{no}} & FFF\textsubscript{mean} & FFF\textsubscript{one} & STD   & SSP   & ASV   & UAP   & EPGD & {\color[HTML]{FE0000}L4A\textsubscript{base}}  & L4A\textsubscript{fuse}  & L4A\textsubscript{ugs} \\ \hline
mIoU   & {\color[HTML]{3531FF} 40.63}    & 40.79     & 40.86       & 42.67 & 41.99 & 41.84 & 41.89 & 41.07 & {\color[HTML]{FE0000} \textbf{39.38}} & 39.59 & 39.44 \\ \hline
\end{tabular}
\end{table}
From the table, we can see that our methods generalize well to the segmentation task. While FFF\textsubscript{no} achieve a mIoU of 40.63\%, that of L4A\textsubscript{no} is 39.38\%.
All the experiments above show the great cross-task transfer-ability of our methods.

\section{Gradient alignment}
\subsection{Gradient alignment: Proof}\label{app:proof}
In this section, we formulate the definition of the gradient alignment and give a brief proof.

\subsubsection{Preliminaries}

Given a convolutional layer $Conv$ with kernel size = $ks$, stride = 1, bias = 0, an input image $\vect{im} \in \mathcal{R}^{in\times in}$, then the output $ReLu\big[Conv(\vect{im},kernel)\big] \in \mathcal{R}^{(in-ks+1)\times (in-ks+1)}$. Note that $ks << in$.

According to the methods for calculating convolution in computers, the input image $\vect{x}$ will be flatten into a vector $\vect{x} \in \mathcal{R}^{n}$, where $n = in \times in$ and the weights of the convolution layer can be reshaped into a matrix $\vect{W} \in \mathcal{R}^{m\times n}$, where $m = (in-ks+1)\times (in-ks+1)$. Then we have the output $\vect{y} = ReLu\big[\vect{W}\vect{x}\big]$.

Denote $\vect{w}_i$ as the i-th row of the matrix $\vect{W}$. Let the elements of the kernel and $\vect{x}$ subject to the standard normal distribution independently.

\begin{lemma}\label{lemma1}
$\mathbb{E}\big[\vect{w}_i\vect{w}_j^{T}\big] = ks^{2}\delta_{i,j}$.
\end{lemma}

\begin{proof}
When $i = j$, we have $\vect{w}_i\vect{w}_i^{T} \sim \mathcal{X}^2(ks^2)$. Thus $\mathbb{E}\big[\vect{w}_i\vect{w}_i^{T}\big] = ks^{2}$.\\
When $i \neq j$, due to the arrangement of the none-zero elements in the matrix $\mathcal{W}$, we have $\vect{w}_i\vect{w}_j^{T} = \sum\limits_{k=1}^{N}x_{k_1}x_{k_2}$, where $x_{k_1},x_{k_2} \sim N(0,1)$ independently and $0 \leq N \leq ks^{2}$. Thus $\mathbb{E}\big[\vect{w}_i\vect{w}_j^{T}\big] = \sum\limits_{k=1}^{N}\mathbb{E}\big[x_{k_1}x_{k_2}\big] = \sum\limits_{k=1}^{N}\mathbb{E}\big[x_{k_1}\big]\mathbb{E}\big[x_{k_2}\big]=0$
\end{proof}

\begin{lemma}\label{lemma2}
$\mathcal{P}\big(\vect{w}_i\vect{w}_j^{T}=0\big) = \frac{\tbinom{n-ks^2}{ks^2}}{\tbinom{n}{ks^2}}$.
\end{lemma}

\begin{proof}
There are only $ks^{2}$ non-zero elements in $\vect{w}_i$. Then we have ${\tbinom{n-ks^2}{ks^2}}$ ways to choose $ks^{2}$ zero elements from $\vect{w}_j$, making the sum of the product zero. Meanwhile, we have ${\tbinom{n}{ks^2}}$ ways to choose $ks^{2}$ elements from $\vect{w}_j$. Finally the probability is the ratio of the two values.
\end{proof}

\begin{assumption}\label{ass}
 $\vect{w}_i\vect{w}_j^{T} = 0$, for $i \neq j $.
\end{assumption}

According to Lemma~\ref{lemma1}, $\mathbb{E}\big[\vect{w}_i\vect{w}_j^{T}\big] = 0$ for $i \neq j$.  According to Lemma~\ref{lemma2}, $\lim\limits_{\frac{ks^2}{n} \to 0}\mathcal{P}\big(\vect{w}_i\vect{w}_j^{T}=0\big)=1$.

\begin{assumption}\label{ass2}
 The elements of $\vect{x_1}$, $\vect{x_2}$ and the kernel subject to the standard normal distribution independently.
\end{assumption}

\subsubsection{Proof}

Let $\vect{x}_1$ and $\vect{x}_2 \in \mathcal{R}^{n}$ be two flattened vectors and let $\vect{W}$ be the weight matrix of the first convolution layer.

For the first iteration of the L4A algorithm, the output of the first convolution layer $\vect{y}_1 = ReLu\big(\vect{W}\vect{x}_1\big)$.
Then the gradient of the loss function in the first step is:

\begin{equation}
\begin{split}
\frac{\partial L}{\partial \vect{x}_1} &= \frac{1}{m} \frac{\partial \vect{y}_1^T\vect{y}_1}{\partial \vect{x}_1} = \frac{1}{m}\sum\limits_{i=1}^{m}\frac{\partial ReLu^{2}(\vect{w}_i \vect{x}_1)}{\vect{x}_1}\vect{w}_i \\&= \frac{2}{m}\sum\limits_{i=1}^{m}ReLu\big(\vect{w}_i\vect{x}_1\big)U\big(\vect{w}_i\vect{x}_1\big)\vect{w}_i =\frac{2}{m}\sum\limits_{i=1}^{m}ReLu\big(\vect{w}_i\vect{x}_1\big)\vect{w}_i 
\end{split}
\end{equation}
where $U(\vect{\cdot})$ denotes the step function.

Considering the update of $\vect{\delta}$, we have the output in the second step $\vect{y}_2 = ReLu\big[\vect{W}\big(\vect{x}_1+\vect{\alpha}\frac{\partial L}{\partial \vect{x}_1}\big)\big]$
where $\vect{}\alpha$ denotes the step size.
Then the gradient of the loss function in the second step is:

\begin{equation}
\begin{split}
\frac{\partial L}{\partial \vect{x}_2} 
&= \frac{1}{m} \frac{\partial \vect{y}_2^T\vect{y}_2}{\partial \vect{x}_2} = \frac{1}{m}\sum\limits_{i=1}^{m}\frac{\partial ReLu^{2}(\vect{w}_i\vect{x}_2+\vect{w}_i\frac{2\alpha}{m}\sum\limits_{j=1}^{m}ReLu\big(\vect{w}_j\vect{x}_1\big)\vect{w}_j^{T})}{\partial\vect{x}_2} \\
& = \frac{1}{m}\sum\limits_{i=1}^{m}\frac{\partial ReLu^{2}(\vect{w}_i\vect{x}_2+\frac{2\alpha}{m}ReLu\big(\vect{w}_i\vect{x}\big)\vect{w}_i\vect{w}_i^{T})}{\partial\vect{x}_2} \\
& = \frac{2}{m}\sum\limits_{i=1}^{m}ReLu\big(\vect{w}_i\vect{x}_2+\frac{2\alpha}{m}ReLu\big(\vect{w}_i\vect{x}\big)\vect{w}_i\vect{w}_i^{T}\big)\vect{w}_i
\end{split}
\end{equation}

Ignoring the update of $\vect{\delta}$, we have the output of the convolution layer in the second step $\vect{y}_2^{*} = ReLu\big(\vect{W}\vect{x}_2\big)$

Then the gradient in the second step can be formulated as:
\begin{equation}
\begin{split}
\frac{\partial L^{*}}{\partial \vect{x}_2} &= \frac{1}{m} \frac{\partial {\vect{y}_2^{*}}^T\vect{y}_2^{*}}{\partial \vect{x}_2} = \frac{1}{m}\sum\limits_{i=1}^{m}\frac{\partial ReLu^{2}(\vect{w}_i \vect{x}_2)}{\vect{x}_2}\vect{w}_i \\&= 2\frac{1}{m}\sum\limits_{i=1}^{m}ReLu\big(\vect{w}_i\vect{x}_2\big)U\big(\vect{w}_i\vect{x}_2\big)\vect{w}_i =2\frac{1}{m}\sum\limits_{i=1}^{m}ReLu\big(\vect{w}_i\vect{x}_2\big)\vect{w}_i 
\end{split}
\end{equation}

\begin{definition}[Gradient alignment]
$\mathcal{GA}=\mathbb{E}_{\vect{x}_1,\vect{x}_2}\big[\frac{\partial L}{\partial \vect{x}_2} \frac{\partial L}{\partial \vect{x}_1}^{T} \big]$
\end{definition} 

Here the gradient alignment measures the similarity between steps of different iterations.

\begin{definition}[Pseudo-gradient alignment]
$\mathcal{PGA}=\mathbb{E}_{\vect{x}_1,\vect{x}_2}\big[\frac{\partial L^{*}}{\partial \vect{x}_2} \frac{\partial L}{\partial \vect{x}_1}^{T} \big]$
\end{definition}

Here the pseudo-gradient alignment shows the similarity when ignoring the update and provides the reference value for easy comparison.

\begin{theorem}
The $\mathcal{GA}$ of L4A is never smaller than the $\mathcal{PGA}$.
\end{theorem}


\begin{proof}
\begin{equation}
\begin{split}
&\mathbb{E}\big(\frac{\partial L}{\partial \vect{x}_2} \cdot \frac{\partial L}{\partial \vect{x}_1}^{T}\big) - \mathbb{E}\big(\frac{\partial L^{*}}{\partial \vect{x}_2} \cdot \frac{\partial L}{\partial \vect{x}_1}^{T}\big) \\ &=\frac{4}{m^{2}}\mathbb{E}\bigg[\sum\limits_{i=1}^{m}ReLu\big(\vect{w}_i\vect{x}_2+\frac{2\alpha}{m}ReLu\big(\vect{w}_i\vect{x}_1\big)\vect{w}_i\vect{w}_i^{T}\big)\vect{w}_i  \sum\limits_{j=1}^{m}ReLu\big(\vect{w}_j\vect{x}_1\big)\vect{w}_j^{T} \bigg]\\
&-\frac{4}{m^{2}}\mathbb{E}\bigg[\sum\limits_{i=1}^{m}ReLu\big(\vect{w}_i\vect{x}_2\big)\vect{w}_i  \sum\limits_{j=1}^{m}ReLu\big(\vect{w}_j\vect{x}_1\big)\vect{w}_j^{T} \bigg]\\
&=\frac{4}{m^{2}}\mathbb{E}\bigg[\sum\limits_{i=1}^{m}ReLu\big(\vect{w}_i\vect{x}_2+\frac{2\alpha}{m}ReLu\big(\vect{w}_i\vect{x}_1\big)\vect{w}_i\vect{w}_i^{T}\big)\vect{w}_i ReLu\big(\vect{w}_i\vect{x}_1\big)\vect{w}_i^{T} \bigg] \\
&-\frac{4}{m^{2}}\mathbb{E}\bigg[\sum\limits_{i=1}^{m}ReLu\big(\vect{w}_i\vect{x}_2\big)\vect{w}_i ReLu\big(\vect{w}_i\vect{x}_1\big)\vect{w}_i^{T} \bigg] \\
&=\frac{4}{m^{2}}\sum\limits_{i=1}^{m}\mathbb{E}\bigg[ \bigg(ReLu\big(\vect{w}_i\vect{x}_2+\frac{2\alpha}{m}ReLu\big(\vect{w}_i\vect{x}_2\big)\vect{w}_i\vect{w}_i^{T}\big)-ReLu\big(\vect{w}_i\vect{x}_2\big) \bigg)ReLu\big(\vect{w}_i\vect{x}_1\big)\vect{w}_i\vect{w}_i^{T} \bigg] \geq 0
\end{split}
\nonumber 
\end{equation}
Note that $ReLu\big(\vect{w}_i\vect{x}_2\big)\vect{w}_i\vect{w}_i^{T} \geq 0$,
thus $ReLu\big(\vect{w}_i\vect{x}_2+\frac{2\alpha}{m}ReLu\big(\vect{w}_i\vect{x}_2\big)\vect{w}_i\vect{w}_i^{T}\big)-ReLu\big(\vect{w}_i\vect{x}_2\big) \geq 0$
\end{proof}

\subsection{Gradient alignment: Simulation}\label{app:gradient}
In this subsection, we provide details about the simulation evaluating the gradient alignment of different algorithms.
First, we run the targeted algorithm 256 times and record the gradients obtained in the algorithm. Then we compute the cosine similarity matrix of the 256 gradients and exclude diagonal elements. Finally, we refer to the average over the similarity matrix as the gradient alignment of the method. Here we provide additional simulation results on Resnet50 and ViT16 in Table~\ref{ap:garesnet50} and Table~\ref{ap:gaViT16} respectively. Note that we report the attack success rates (\%).

\makeatletter\def\@captype{table}\makeatother
\begin{minipage}{0.5\textwidth}\
\centering
\setlength{\tabcolsep}{1.5mm}
\begin{tabular}{l|ccc}
\hline
\textbf{Resnet50} & ${\mathcal{GA}}$     & ImageNet & AVG     \\ \hline
FFF\textsubscript{mean}      & 0.0158 & 45.98  & 52.10 \\
DR       & 0.0449 & 45.46  & 48.26 \\
UAP      & 0.0020 & 95.34  & 55.16 \\
UAPEPGD  & 0.0010 & 93.67  & 69.28 \\
SSP      & 0.0449 & 44.28  & 53.64 \\
L4A\textsubscript{base}  & 0.5489 & 45.54  & 71.16 \\ \hline
\end{tabular}
\caption{Gradient alignment on Resnet50}\label{ap:garesnet50}
\end{minipage}
\makeatletter\def\@captype{table}\makeatother
\begin{minipage}{0.45\textwidth}
\centering
\setlength{\tabcolsep}{1.5mm}
\begin{tabular}{l|ccc}
\hline
\textbf{ViT16}   & GA     & ImageNet & AVG     \\ \hline
FFF\textsubscript{mean}     & 0.1005 & 99.88  & 71.06 \\
DR      & 0.0861 & 56.06  & 27.02 \\
UAP     & 0.0504 & 98.46  & 55.16 \\
UAPEPGD & 0.0049 & 97.66  & 66.95 \\
SSP     & 0.1279 & 80.63  & 53.40 \\
L4A\textsubscript{base} & 0.1386 & 94.15  & 94.00 \\ \hline
\end{tabular}
\caption{Gradient alignment on ViT16}\label{ap:gaViT16}
\end{minipage}

\section{Ablation studies}
\subsection{Effect of fusing the knowledge of different layers}
\begin{figure}[H]
\centering  
\subfigure[Resnet101]{
\label{fig5.1}
\includegraphics[width=0.32\textwidth]{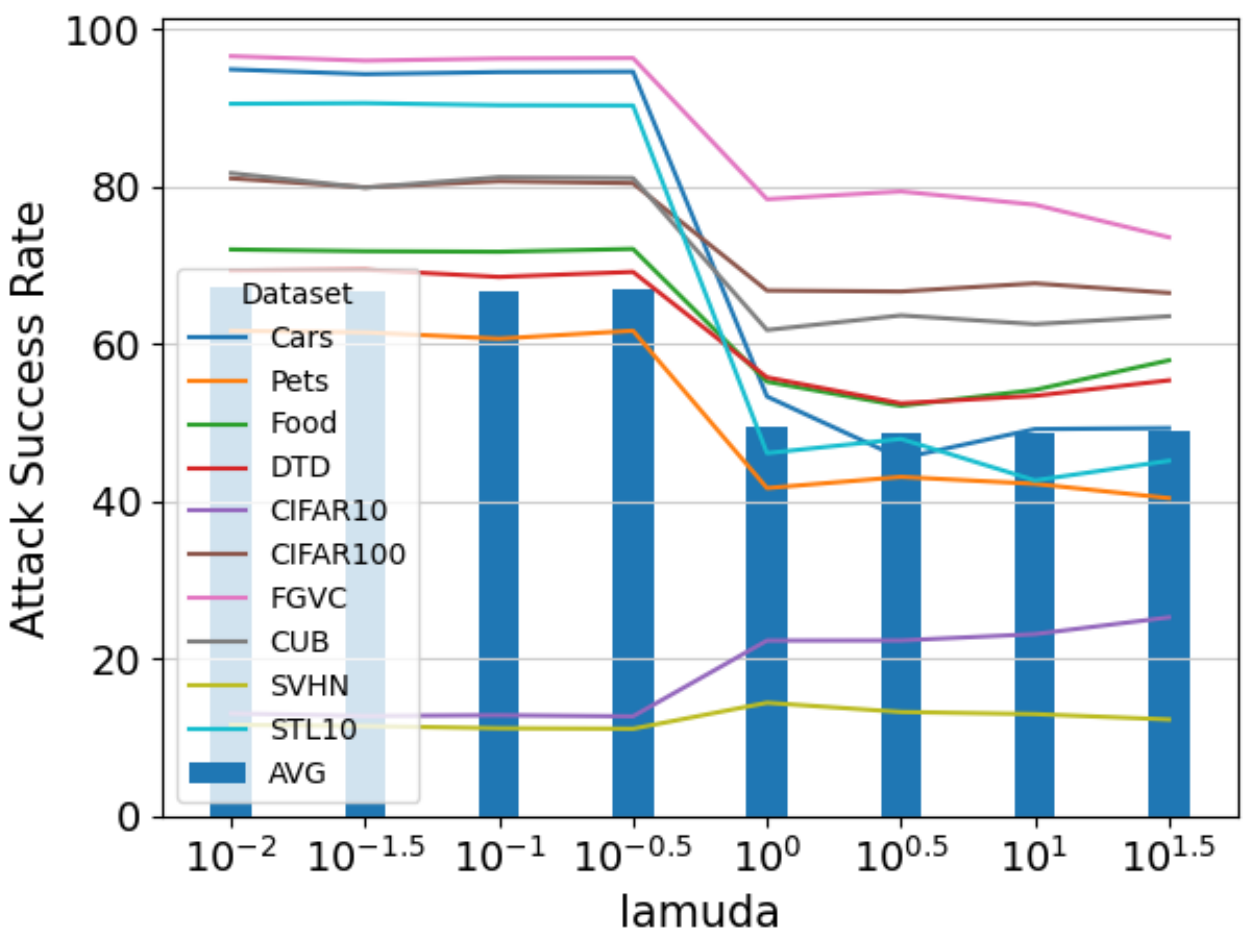}}
\subfigure[Resnet50]{
\label{fig5.2}
\includegraphics[width=0.32\textwidth]{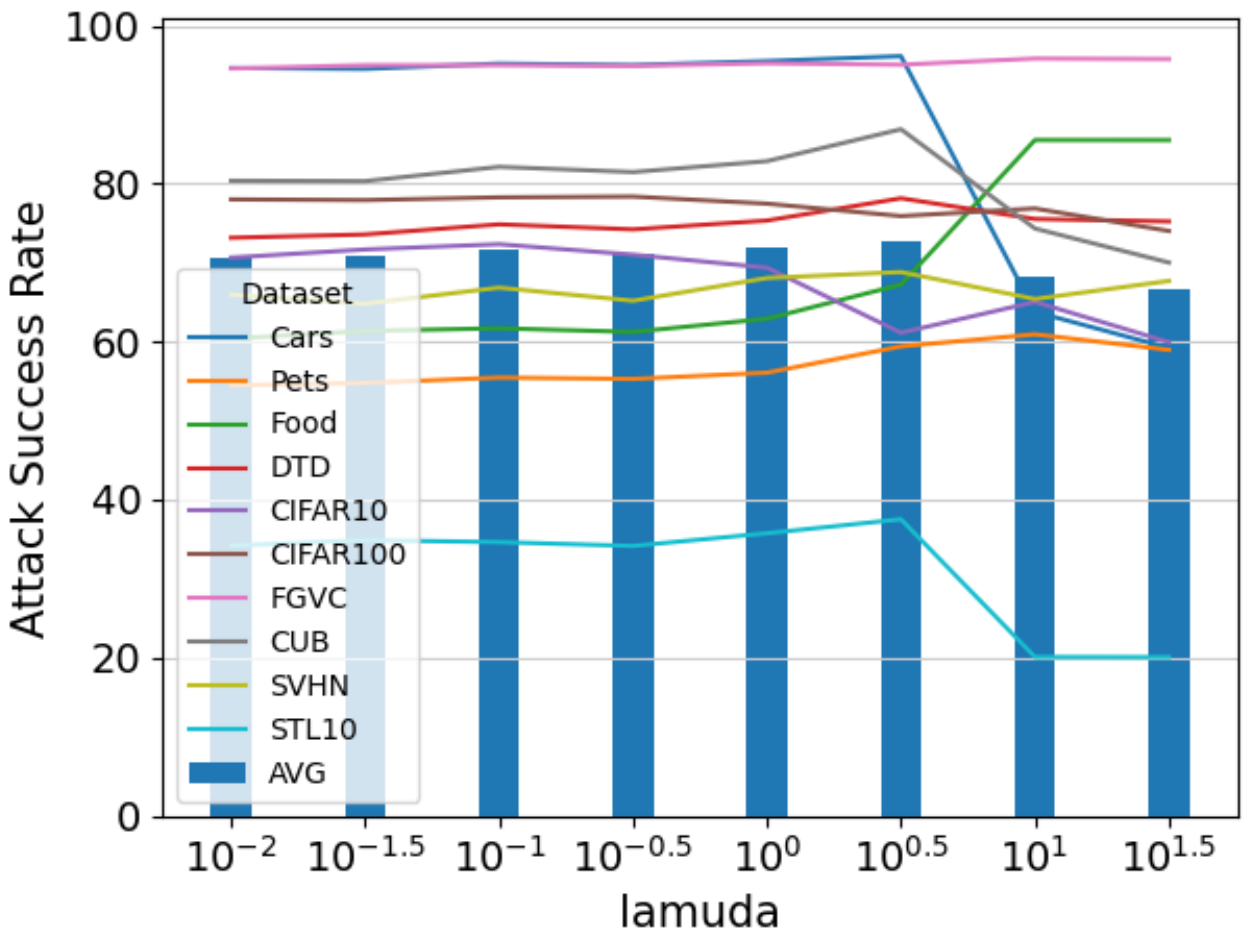}}
\subfigure[ViT]{
\label{fig5.3}
\includegraphics[width=0.32\textwidth]{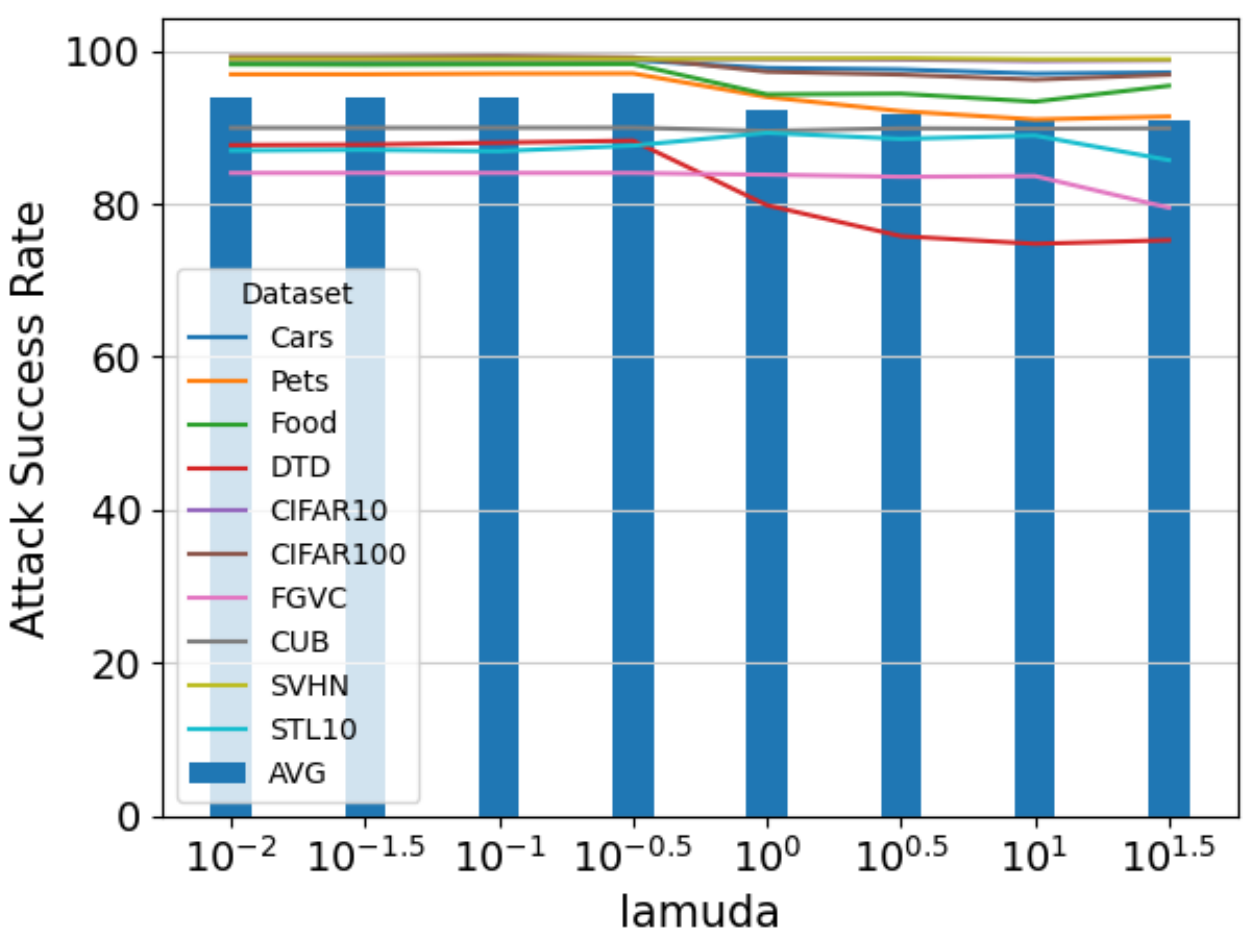}}
\caption{The effect of the scale factor in L4A\textsubscript{fuse}} 
\label{fig:lamuda_fuse}
\end{figure}

\vspace{-1ex}
Here we discuss the effect of the scale factor $\vect{\lambda}$ to fuse the knowledge from different layers, and the results are shown in Fig~\ref{fig:lamuda_fuse}. For Resnet50, in Fig.~\ref{fig5.2}, setting $\vect{\lambda}$ as $10^{0.5}$ can boost the performance by 1.5\%.
\subsection{Effect of using high-level loss}
To study the effect of utilizing the high-level features, we choose Resnet50 pre-trained by SimCLRv2 as the target. In previous experiments, we found that the $L4A_{\text{fuse}}$ method performs best with $\lambda=1$. Thus, we fix it and add a new loss term summing over the lifting loss of the third, fourth and fifth layers, which is balanced by a hyperparameter $\mu$ (Note that we divided the Resnet50 into five blocks, meaning that it has five layers in total in our settings. Please refer to Appendix~\ref{modelarchi} for more details about the model architecture). 
Finally the training loss of the experiments testing high-level layers can be formulated as follows:
\begin{equation}
\min_{\delta}  L(f_{\theta},x,\delta) = -\mathbb{E}_{x\sim D_p}\bigg[\sum_{i=1}^{2}||f_{\theta}^{i}(x+\delta)||_F^{2}+\mu\sum_{j=3}^{5}||f_{\theta}^{j}(x+\delta)||_F^{2}\bigg],
\end{equation}
We report the attack success rate (\%) of using different $\mu$ against Resnet50 pre-trained by SimCLRv2. Note that C10 stands for CIFAR10, and C100 stands for CIFAR100. Results are shown in the Table~\ref{mu}.
\begin{table}[H]\footnotesize
\setlength{\tabcolsep}{5pt}
\caption{The attack success rate(\%) of different $\mu$ against \textbf{Resnet50} pretrained by \textbf{SimCLRv2}. Note that C10 stands for CIFAR10 and C100 stands for CIFAR100.}
\begin{tabular}{l||ccccccc|ccc|c}
\hline
$\vect{\mu}$
& Cars                                    & Pets                                    & Food                           & DTD                                     & FGVC                                    & CUB                                     & SVHN                                    & C10                                 & C100                                & STL10                                   & AVG                                     \\ \hline\hline
0       & 96.00          & \textbf{59.80} & \textbf{65.00} & \textbf{77.93} & 95.02          & \textbf{85.05} & 69.39          & 64.41          & 76.29          & 37.54          & \textbf{72.64} \\
0.01    & 95.32          & 56.99          & 63.60          & 76.06          & 94.93          & 83.31          & \textbf{69.65} & 67.30          & 77.55          & \textbf{37.76} & 72.25          \\
0.1     & 95.00          & 56.06          & 62.02          & 75.00          & \textbf{95.32} & 82.21          & 66.76          & 70.36          & \textbf{77.86} & 35.28          & 71.59          \\
1       & 95.16          & 56.55          & 62.11          & 74.95          & 95.20          & 82.27          & 67.15          & 69.20          & 78.47          & 35.75          & 71.68          \\
10      & \textbf{96.17} & 55.16          & 62.08          & 77.82          & 95.26          & 83.52          & 66.48          & 67.37          & 77.56          & 33.39          & 71.48          \\
100     & 61.78          & 42.71          & 59.86          & 70.05          & 86.68          & 66.26          & 63.43          & \textbf{86.99} & 64.92          & 21.78          & 62.44          \\ \hline
\end{tabular}\label{mu}
\end{table}

As seen from the last column, the larger the weight of the loss of the high-level layers, the worse it performs. Moreover, when the loss of high-level layers overwhelms the low-level ones, the method suffers a significant performance drop (over 10\%) in attack success rates. These results show that adding the high-level loss to the training loss bears negative effects.
\subsection{Hyperparameters in the Uniform Gaussian Sampling.}

We chose these hyperparameters as $\mu_l=0.4$, $\mu_h=0.6$, $\sigma_l=0.05$, $\sigma_h=0.1$ in the experiments. The reasons are as follows. For $\mu_l$ and $\mu_h$, we aim to make the mean $\mu$ drawn from $U(\mu_l,\mu_h)$ distributed around 0.5, since the input images are normalized to $[0,1]$. Thus we tried several configurations of $(\mu_l,\mu_h)$, such as (0.4, 0.6) and (0.45, 0.55), and found that (0.4, 0.6) performs best. 
For $\sigma_{l}$ and $\sigma_{h}$, we hope that most of the samples $n_0\sim N(\mu, \sigma)$ lie in $[0,1]$. Thus $\sigma\sim U(\sigma_l,\sigma_h)$ cannot be too large. We also tried some configurations of ($\sigma_{l}$, $\sigma_{h}$) and found that (0.05, 0.1) performs best.

Interestingly, the set (0.4, 0.6, 0.05, 0.10) generalizes well across the three models. Thus we did not tune these hyperparameters for each model but adopted a single configuration in Table~\ref{tab:res101}, Table~\ref{tab:res50}, and Table~\ref{tab:vit}.

\subsection{Pixel-level perturbations}
To test the pixel-level perturbations, we add $\epsilon=0.05$ to the input images and then evaluate the performance on the three pre-trained models studied in the paper. Then we report the average attack success rate(\%) on the ten datasets in Table~\ref{tab:pixcom}, and detailed performance in Table~\ref{tab:pixde}. Note that SimR101, SimR101 and MAEViT stand for Resnet101 pretrained by SimCLRv2, Resnet50 pretrained by SimCLRv2 and ViT-base-16 pretrained by MAE, respectively.

\begin{table}[H]\footnotesize
\setlength{\tabcolsep}{3.5pt}
\caption{The average attack success of different methods against the three models. Note that C10 stands for CIFAR10 and C100 stands for CIFAR100.}
\begin{tabular}{l||cccccccc|ccc|c}
\hline
methods       & FFF\textsubscript{no} & FFF\textsubscript{mean} & FFF\textsubscript{one} & STD   & SSP   & ASV   & UAP   & EPGD & L4A\textsubscript{base}  & L4A\textsubscript{fuse}  & L4A\textsubscript{ugs} & Pixel\\ \hline\hline
SimR101 & 48.55 & 44.22   & 40.26  & 42.63 & 40.75 & 46.65 & 43.86 & 59.34   & 66.89   & 71.90   & 72.20  & 12.97    \\
SimR50  & 43.86 & 52.10   & 52.98  & 48.26 & 53.64 & 58.19 & 55.16 & 69.28   & 71.16   & 72.64   & 77.80  & 13.76    \\
MAEViT  & 77.69 & 71.06   & 74.35  & 27.02 & 53.40 & 22.64 & 55.16 & 66.95   & 94.00   & 94.42   & 95.30  & 12.52    \\ \hline
\end{tabular}\label{tab:pixcom}
\end{table}

\begin{table}[H]\footnotesize
\setlength{\tabcolsep}{5pt}
\caption{The attack success rate(\%) of the pixel-level attack on ten datasets. Note that C10 stands for CIFAR10 and C100 stands for CIFAR100.}
\begin{tabular}{l||ccccccc|ccc|c}
\hline
ASR
& Cars                                    & Pets                                    & Food                           & DTD                                     & FGVC                                    & CUB                                     & SVHN                                    & C10                                 & C100                                & STL10                                   & AVG                                     \\ \hline\hline
SimR101 & 10.42 & 9.70  & 12.15 & 29.36 & 23.79 & 21.25 & 2.61 & 2.21    & 15.46    & 2.75  & 12.97 \\
SimR50  & 10.73 & 11.61 & 12.15 & 29.57 & 26.67 & 22.47 & 2.68 & 2.53    & 16.14    & 3.09  & 13.76 \\
MAEViT & 9.92  & 6.90  & 10.53 & 26.28 & 33.75 & 17.96 & 2.64 & 2.50    & 11.99    & 2.76  & 12.52 \\ \hline
\end{tabular}\label{tab:pixde}
\end{table}
As we can see from the tables, the pixel-level perturbations have little effect on the predictions.

\section{Datasets}\label{appdataset}
We evaluate the performance of pre-trained adversarial perturbations on the CIFAR100 and CIFAR10~\cite{krizhevsky2009cifar100}, STL10~\cite{coates2011stl10}, Cars~\cite{krause2013cars}, Pets~\cite{parkhi2012pets}, Food~\cite{bossard2014food}, DTD~\cite{cimpoi2014DTD}, FGVC~\cite{maji2013fgvc}, CUB~\cite{wah2011cub}, SVHN~\cite{netzer2011svhn}. We report the calibration (fine-grained or coarse-grained) and the accuracy on clean samples in Table~\ref{ap:dataset}.

\begin{table}[h]\tiny
\caption{Calibration and ACC (\%)}
\begin{tabular}{l||ccccccc|ccc}
\hline
Dataset       & Cars    & Pets    & Food    & DTD     & FGVC    & CUB     & SVHN    & CIFAR10 & CIFAR100 & STL10   \\ \hline\hline
Calibration   & fine    & fine    & fine    & fine    & fine    & fine    & fine    & coarse  & coarse   & coarse  \\
Resnet101 ACC & 89.80 & 90.60 & 87.90 & 71.01 & 77.04 & 78.78 & 97.40 & 97.85 & 84.81  & 97.33 \\
Resnet50 ACC  & 89.35 & 88.20 & 87.84 & 70.60 & 74.01 & 78.13 & 97.40  & 97.51 & 84.03  & 97.00 \\
ViT ACC       & 90.03 & 93.48 & 89.60 & 73.60 & 67.16 & 82.32 & 97.38 & 98.10 & 88.03  & 97.20 \\ \hline
\end{tabular}
\end{table}\label{ap:dataset}

Our datasets do not involve these issues.

\section{Visualisation of Perturbations}
\begin{figure}[H] 
\centering 
\includegraphics[width=1\textwidth]{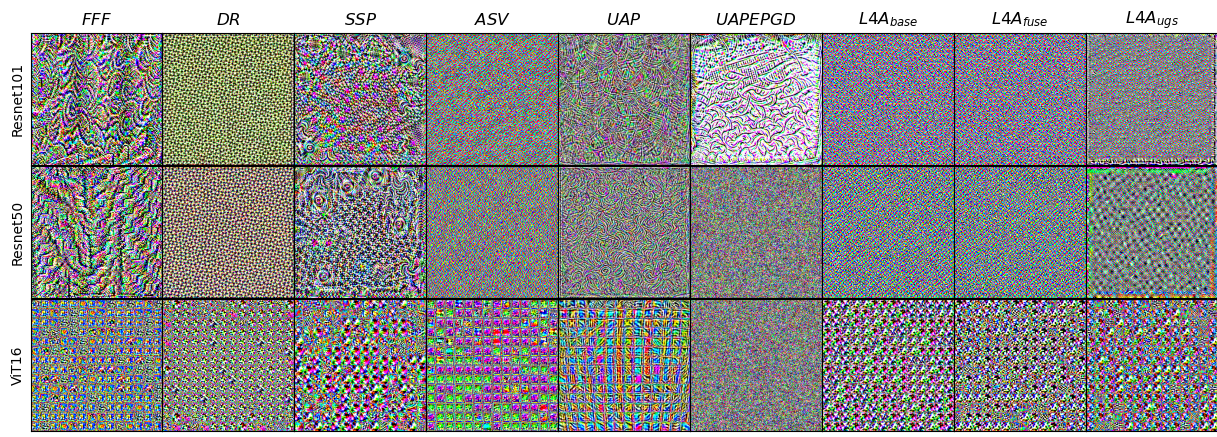} 
\caption{Visualization of pre-trained adversarial perturbations} 
\label{Fig.main2} 
\end{figure}
\section{Implementations}
\subsection{Model architecture}\label{modelarchi}
To evaluate the effect of attacking different layers, we divide Resnet50, Resnet101 and ViT16 into 5 parts. Here we provide the mapping relationship from the original name to the five layers, respectively.

\begin{table}[h]
\caption{Model architecture}
\begin{tabular}{l||ccccc}
\hline
Layers    & layer1        & layer2            & layer3            & layer4            & layer5            \\ \hline
Resnet50  & net{[}0{]}    & net{[}1{]}        & net{[}2{]}        & net{[}3{]}        & net{[}4{]}        \\
Resnet101 & net{[}0{]}    & net{[}1{]}        & net{[}2{]}        & net{[}3{]}        & net{[}4{]}        \\
ViT16     & blocks{[}0{]} & blocks{[}1,2,3{]} & blocks{[}4,5,6{]} & blocks{[}7,8,9{]} & blocks{[}10,11{]} \\ \hline
\end{tabular}
\end{table}
\subsection{Resources}
We use one Nvidia GeForce RTX 2080 Ti for generating and evaluating PAPs.

\end{document}